%% file: main.tex
\documentclass[10pt,twocolumn,letterpaper]{article}

\usepackage[pagenumbers]{cvpr}
\usepackage{amssymb}
\usepackage{changepage}
\usepackage{pgfplots}
\usepackage{tikz}
\pgfplotsset{compat=newest}

\input{preamble}

\definecolor{cvprblue}{rgb}{0.21,0.49,0.74}
\usepackage[pagebackref,breaklinks,colorlinks,allcolors=cvprblue]{hyperref}

\def\paperID{} 
\def\confName{CVPR}
\def\confYear{2026}

\title{\oursx: Scaling Bimanual Motion and Interaction Generation}

\author{Zimu Zhang$^{1\dag}$ \quad Yucheng Zhang$^{1\dag}$ \quad Xiyan Xu$^1$ \quad Ziyin Wang$^1$ \quad Sirui Xu$^{1\ddag}$ \quad
Kai Zhou$^{2,3}$ \\ \quad Bing Zhou$^3$ \quad Chuan Guo$^3$ \quad Jian Wang$^3$ \quad Yu-Xiong Wang$^{1}$ \quad
Liang-Yan Gui$^{1}$\\
\small{$^{1}$University of Illinois Urbana-Champaign \quad $^{2}$Specs Inc. \quad $^{3}$Snap Inc.}\\
\small{$^{\dag}$Equal Contribution \quad $^{\ddag}$Project Lead}\\
\small\url{https://handx-project.github.io}}
\begin{document}

\twocolumn[{%
   \renewcommand\twocolumn[1][]{#1}
   \maketitle 
   \vspace{-0.8cm}
   \begin{center}
      \centering
      
      \includegraphics[width=\textwidth]{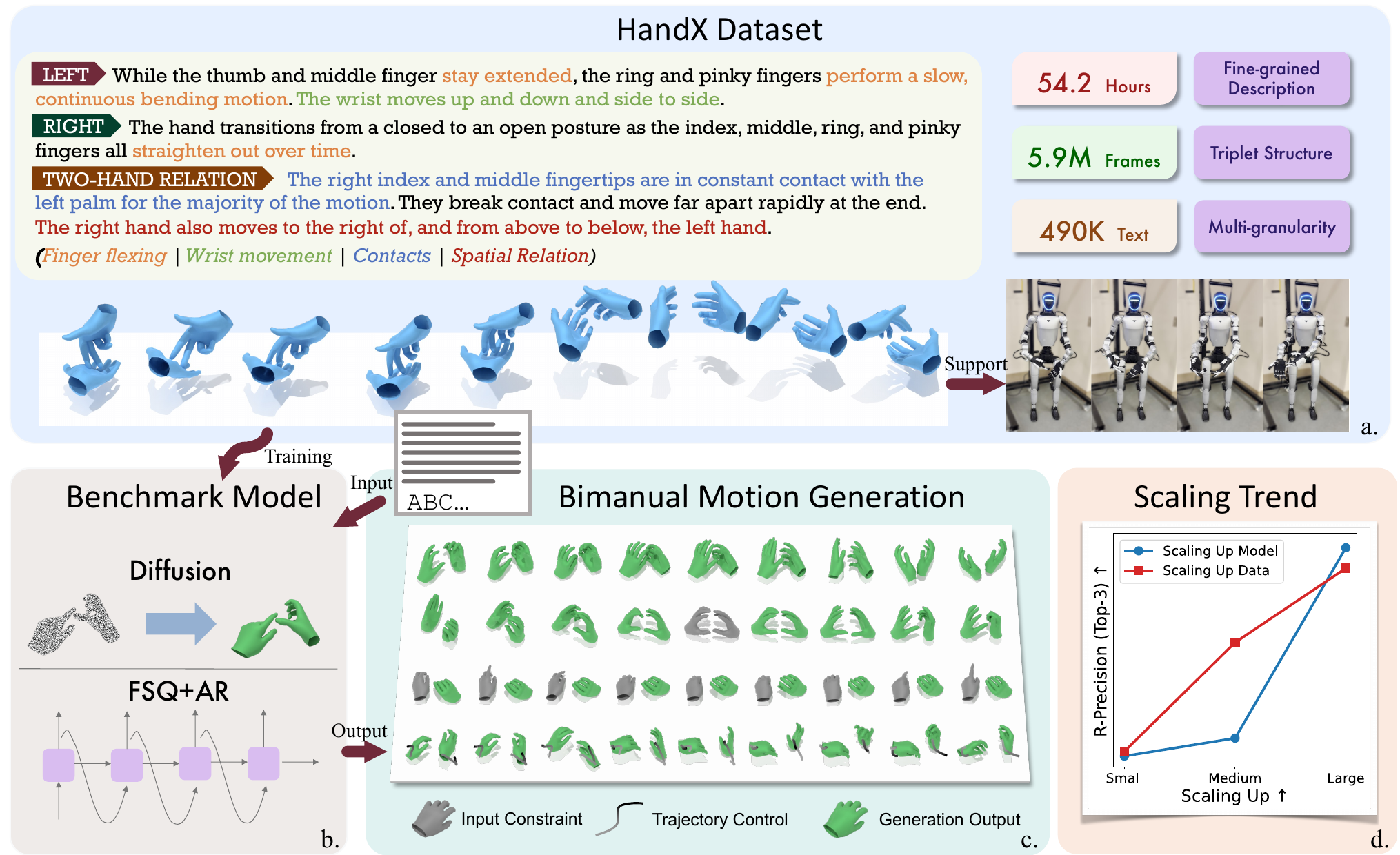} 
      \vspace{-0.2cm} 
      \captionof{figure}{\textbf{(a)} We introduce \textbf{\oursx}, a large-scale dataset of \textit{bimanual} and \textit{dexterous} motions paired with \textit{fine-grained} textual descriptions. The examples highlight the high-fidelity captures produced by our motion capture system (Figure~\ref{fig:mocap_setup}), and demonstrate instantiation on a real-world humanoid with dexterous hands.
      \textbf{(b)} We benchmark two generative paradigms: diffusion-based and autoregressive (AR) models. 
      \textbf{(c)} Our models support flexible conditioning and synthesize highly dynamic, expressive hand motions.
      \textbf{(d)} We observe clear scaling trends: increasing dataset size and model capacity yields substantial performance gains.
      }
      \label{fig:teaser}
   \end{center}
   \vspace{0.5cm} 
}]

\input{sec/0_abstract}    
\input{sec/1_intro}
\input{sec/related_work}
\input{sec/3_Dataset_v6}
\input{sec/4_auto_annotation_v5}
\input{sec/5_method_v2}

\input{sec/experiments} 
\input{sec/8_conclusion}

{
    \small
    \bibliographystyle{ieeenat_fullname}
    \bibliography{main}
}

\appendix

\input{sec/X_suppl}

\end{document}

%% file: preamble.tex
\usepackage{xspace}
\usepackage{soul}
\usepackage{calc}
\newcommand{\oursx}{HandX\xspace}
\usepackage{float}
\usepackage[table]{xcolor}
\usepackage{booktabs}

\usepackage{makecell}
\usepackage{pifont}
\definecolor{oursbg}{RGB}{235,245,255}
\definecolor{groupgray}{RGB}{245,245,245}

\usepackage{multirow}

%% file: sec/0_abstract.tex
\begin{abstract}
Synthesizing human motion has advanced rapidly, yet realistic hand motion and bimanual interaction remain underexplored. Whole-body models often miss the fine-grained cues that drive dexterous behavior, finger articulation, contact timing, and inter-hand coordination, and existing resources lack high-fidelity bimanual sequences that capture nuanced finger dynamics and collaboration. To fill this gap, we present HandX, a unified foundation spanning data, annotation, and evaluation. We consolidate and filter existing datasets for quality, and collect a new motion-capture dataset targeting underrepresented bimanual interactions with detailed finger dynamics. For scalable annotation, we introduce a decoupled strategy that extracts representative motion features, e.g., contact events and finger flexion, and then leverages reasoning from large language models to produce fine-grained, semantically rich descriptions aligned with these features. Building on the resulting data and annotations, we benchmark diffusion and autoregressive models with versatile conditioning modes. Experiments demonstrate high-quality dexterous motion generation, supported by our newly proposed hand-focused metrics. We further observe clear scaling trends: larger models trained on larger, higher-quality datasets produce more semantically coherent bimanual motion. Our dataset is released to support future research.
\end{abstract}

%% file: sec/1_intro.tex
\section{Introduction}
\label{sec:intro}
Natural communication and skilled manipulation rely heavily on the hands. Despite impressive advances in human animation~\cite{tevet2022human}, human-object interaction~\cite{xu2026interprior}, and video generation~\cite{shao2024human4dit}, most methods still treat hands as an afterthought. As a result, they often miss the fine-grained cues that make hand motion both believable and functional, including precise finger articulation, well-timed contact, and smooth bimanual coordination under semantic intent. These limitations hinder deployment in immersive media, telepresence, embodied AI, and human-computer interaction, where realistic hand motion is essential.

A key bottleneck is the lack of suitable data and an established evaluation protocol. Most human motion and interaction datasets~\cite{guo2022generating, xu2025interact} emphasize locomotion and loco-manipulation but provide limited hand detail, while hand-centric datasets~\cite{zhang2024both2hands,lin2025handdiffuse,moon2020interhand2,kwon2021h2o,fu2025gigahands} focus narrowly on object interaction, miss fine-grained finger dynamics, or use coarse annotations. In addition, mismatched skeletons, frame rates, and annotation protocols hinder unifying data across sources. Finally, existing metrics rarely evaluate hand fidelity or bimanual coordination, making it hard to diagnose failures and measure progress.

To tackle these challenges, we build a unified data foundation for bimanual motion generation, which we call \textbf{HandX}. We consolidate large egocentric and human-object interaction datasets into a standardized corpus with strict quality control (Figure~\ref{fig:teaser}), converting all sequences to a shared representation and filtering implausible or inactive segments. Even after consolidation, a key gap remains: existing data lack high-fidelity bimanual motion that captures fine finger coordination and contact dynamics. We therefore collect a complementary motion-capture dataset of dexterous two-hand interactions (Figure~\ref{fig:mocap_setup}). To scale semantic annotations over all these data, we propose a two-stage strategy that decouples motion understanding from language generation: we first extract structured event descriptors, \eg, touch, slide, and release, then leverage large language model (LLM) reasoning to produce fine-grained descriptions aligned with these events. This enables scalable, consistent annotation with minimal manual effort.

Building on {HandX}, we benchmark two representative paradigms for hand-centric motion generation: a diffusion-based model and an autoregressive, token-based model. To increase versatility, we leverage masked conditioning so a single model supports diverse control modes, including hand reaction generation, motion in-betweening, and keyframe-guided synthesis. We additionally introduce contact-focused metrics to evaluate interaction fidelity. Crucially, we exploit {HandX} to study scaling behavior: in our core text-to-motion benchmark, increasing model capacity and training data consistently improves text alignment and contact accuracy. We further demonstrate that the learned dexterous skills transfer to a humanoid platform equipped with dexterous robot hands, as shown in Figure~\ref{fig:teaser}.

In summary, we establish a unified framework for bimanual motion and interaction generation. We (\textbf{a}) build a hand-centric corpus by consolidating large-scale datasets, and complement it with a new motion-capture dataset emphasizing dexterous two-hand interactions; (\textbf{b}) develop a scalable annotation strategy that produces structured, fine-grained descriptions via feature extraction and LLM reasoning; and (\textbf{c}) benchmark diffusion and autoregressive models with scaling trend analysis on model and data sizes. These contributions provide a foundation for future research on expressive hand motion and interaction synthesis.

%% file: sec/related_work.tex
\section{Related Work}
\noindent\textbf{Human Motion Generation.}
Human motion generation has evolved through several stages. Early work uses latent-variable models and recurrent architectures to map language to motion sequences~\cite{petrovich22temos,guo2022generating,ahn2018text2action,ahuja2019language2pose}. Later methods explore autoregressive generation~\cite{jiang2023motiongpt,xiao2025motionstreamer,zhang2023generating,zou2024parco,lu2025scamo} in parallel with diffusion models~\cite{zhang2022motiondiffuse,tevet2022human,zhang2023remodiffuse,shafir2024human,wang2026unleashing,xu2023stochastic,xu2025moreact,xu2023interdiff,xu2025interact}, emerging as the predominant approaches due to their fidelity and controllability.
Despite this progress, most text-to-motion models do not capture fine-grained hand motion because widely-used datasets~\cite{guo2022generating,xu2025interact} lack articulated hands and instead treat them as rigid end-effectors in SMPL~\cite{SMPL:2015}. Human-object interaction work that includes hand pose and contact~\cite{xu2023interdiff,xu2024interdreamer,wang2026unleashing} typically emphasizes object manipulation, with limited coverage of bimanual coordination and inter-hand contact dynamics. Consequently, current methods remain insufficient for generating realistic, semantically grounded two-hand motion with dexterous contact.

\noindent
\textbf{Hand Motion Generation.}
Hand motion synthesis has been studied under a variety of conditioning modalities. A substantial body of work focuses on audio-driven co-speech gestures~\cite{saleh2022hybrid,abel2024towards,ferstl2019multi,gjaci2022towards,zhu2023taming,yang2023diffusestylegesture,chen2024diffsheg,liu2024emage,liu2025gesturelsm}. Other directions include motion-to-motion generation conditioned on past motion or trajectories~\cite{wen2024generative,lin2025handdiffuse}, body- or object-conditioned motion synthesis and correction~\cite{taheri2024grip,zhang2025bimart,zhang2025manidext,zhou2024gears,wang2024furelise,xu2024synchronize,zhou2022toch,xu2025dexplore}, and vision-based motion forecasting~\cite{liu2022joint,prakash2025bimanual}. Hand motion reconstruction~\cite{fan2024hold,ye2026whole,zhang2025hawor,fu2026egograsp,yu2025dyn} can also be viewed as a form of synthesis. Despite their effectiveness, these methods are not designed to generate hand motion directly from free-form natural language.
Text-driven hand motion synthesis remains relatively underexplored. Recent progress in text-conditioned hand-object interaction adopts diffusion~\cite{cha2024text2hoi,christen2024diffh2o,li2025latenthoi,zhang2025openhoi} or autoregressive models~\cite{huang2025hoigpt}. However, these methods are largely restricted to object-centric settings and offer limited coverage of inter-hand coordination and bimanual contact dynamics. Text-guided gesture and sign-language generation~\cite{baltatzis2024neural,bensabath2025text,fang2025signllm,zuo2025signs} targets communicative motion, prioritizing expressive or linguistic intent over general-purpose motion, and therefore lacks the finger-level dexterity and interaction diversity needed for bimanual synthesis. Concurrently, CLUTCH~\cite{thambiraja2026clutch} generates in-the-wild hand motion from text using an autoregressive model and shows promising coverage of everyday actions, but its action-level input limits motion granularity. Overall, there remains a clear gap in generating fine-grained bimanual hand motion from text, particularly for actions requiring coordinated interaction and contact-aware reasoning.

\noindent\textbf{Hand Motion Datasets.}
The limitations of text-driven models are partially from existing datasets. Full-body motion datasets with articulated hands, such as Motion-X~\cite{lin2023motion} and InterAct~\cite{xu2025interact}, provide textual annotations mainly for whole-body motion rather than fine-grained hands. In contrast, hand-centric datasets often either lack language supervision, such as InterHand2.6M~\cite{moon2020interhand2} and HandDiffuse~\cite{lin2025handdiffuse}, or provide annotations limited to specific domains. A major example is hand-object interaction datasets~\cite{fan2023arctic,kwon2021h2o,liu2022hoi4d,liu2024taco,taheri2020grab,hoque2025egodex}, which are largely object-centric and typically annotated with categorical action labels rather than descriptive, general-purpose text. GigaHands~\cite{fu2025gigahands} offers richer text supervision, but still focuses mainly on object manipulation or predefined gestures, leaving broader bimanual motion and nuanced hand-hand contact underexplored. Sign language datasets~\cite{bilge19zsslr,9681230} also pair text with hand motion, but their data are highly structured and specialized for communication. Recent efforts have begun to scale hand motion data. BOTH2Hands~\cite{zhang2024both2hands} provides 8.31 hours of bimanual motion with finger-level text annotations. Concurrently, BOBSL3DT~\cite{bensabath2025text} builds over 1M motion-text pairs for sign language from monocular reconstruction, while CLUTCH~\cite{thambiraja2026clutch} reconstructs 32K in-the-wild hand motion sequences with annotations by vision-language models. However, BOBSL3DT remains specialized to sign language with limited bimanual interaction, CLUTCH uses action-level descriptions with limited granularity, and both are constrained by monocular reconstruction noise. Overall, existing datasets still lack the precision, diversity, and rich inter-hand contact needed for learning fine-grained bimanual motion from text. HandX is proposed to bridge this gap.

%% file: sec/3_Dataset_v6.tex
\section{Dataset}
\label{sec:dataset}
Most existing motion datasets are not well suited for fine-grained bimanual text-to-motion synthesis, because they lack sufficient hand detail, scale, or interaction richness. To address this gap, we introduce \oursx, a large-scale benchmark for fine-grained bimanual text-to-hand motion generation.
We build \oursx in two steps: (\textbf{a}) \textit{aggregating high-quality open-source data} with bimanual motion~\cite{fu2025gigahands, banerjee2025hot3d, fan2023arctic, kwon2021h2o, wang2023holoassist}, canonicalized into a unified skeletal representation and coordinate system for consistency across heterogeneous sources, while filtering out low-quality sequences; and (\textbf{b}) \textit{capturing high-quality bimanual interaction} with a marker-based optical motion capture system to record dexterous two-hand motion and rich inter-hand contact in natural daily activities.
As shown in Table~\ref{tab:dataset_comparison_new}, \oursx is distinguished by its dynamic and comprehensive collection of \textit{contact-rich} interactions. We further segment all sequences into clips and apply an \textit{intensity-aware} filter based on joint angular velocity, removing dominated static or near-static segments that may cause generative models to \textit{freeze}, and retaining only meaningful interactions, as detailed in Sec.~\ref{sec:Intensity Aware} of the supplementary material.

\begin{table*}[t]
    \centering
    \setlength{\tabcolsep}{5pt}
    \renewcommand{\arraystretch}{1.18}

    \begin{minipage}[c]{0.57\textwidth}
        \centering
        \resizebox{\textwidth}{!}{
        \begin{tabular}{lcccc}
            \toprule
            \rowcolor{groupgray}
            \textbf{Dataset} & \makecell{\textbf{Duration} \\ \textbf{(h)}} & \makecell{\textbf{Frames} \\ \textbf{(M)}} & \textbf{Text Granularity} & \makecell{\textbf{Text} \\ \textbf{(K)}} \\
            \midrule

            Motion-X~\cite{lin2023motion} & 144.2 & 15.6 & coarse & 8.1 \\
            InterAct~\cite{xu2025interact} & 30.7 & 3.3 & coarse & 48.6\\

            \addlinespace[2pt]
            \multicolumn{5}{l}{\textit{Hand motion datasets}} \\
            BOTH2Hands~\cite{zhang2024both2hands} & 8.31 & 1.8 & coarse & 23.5 \\
            HandDiffuse~\cite{lin2025handdiffuse} & 2.0 & 0.25 & -- & -- \\
            InterHand2.6M~\cite{moon2020interhand2} & 24.0 & 2.6 & -- & -- \\

            \addlinespace[2pt]
            \multicolumn{5}{l}{\textit{Hand-object / egocentric datasets}} \\
            GigaHands~\cite{fu2025gigahands} & 2.58 (34.0) & 0.28 (3.7) & coarse & 84 \\
            HOT3D~\cite{banerjee2025hot3d} & 0.44 (3.90) & 0.05 (0.42) & -- & -- \\
            ARCTIC~\cite{fan2023arctic} & 1.06 (2.02) & 0.11 (0.22) & action & -- \\
            H2O~\cite{kwon2021h2o} & 0.47 (1.06) & 0.05 (0.11) & action & -- \\
            HoloAssist~\cite{wang2023holoassist} & 49.3 (161.2) & 5.32 (17.4) & coarse & 1.8 \\

            \midrule
            \rowcolor{oursbg}
            \textbf{\oursx\ (Ours)} & \textbf{54.2} & \textbf{5.9} & \textbf{fine-grained} & \textbf{485.7} \\
            \bottomrule
        \end{tabular}
        }
    \end{minipage}
    \hfill
    \begin{minipage}[c]{0.41\textwidth}
        \vspace{0pt}
        \centering
        \includegraphics[width=\textwidth]{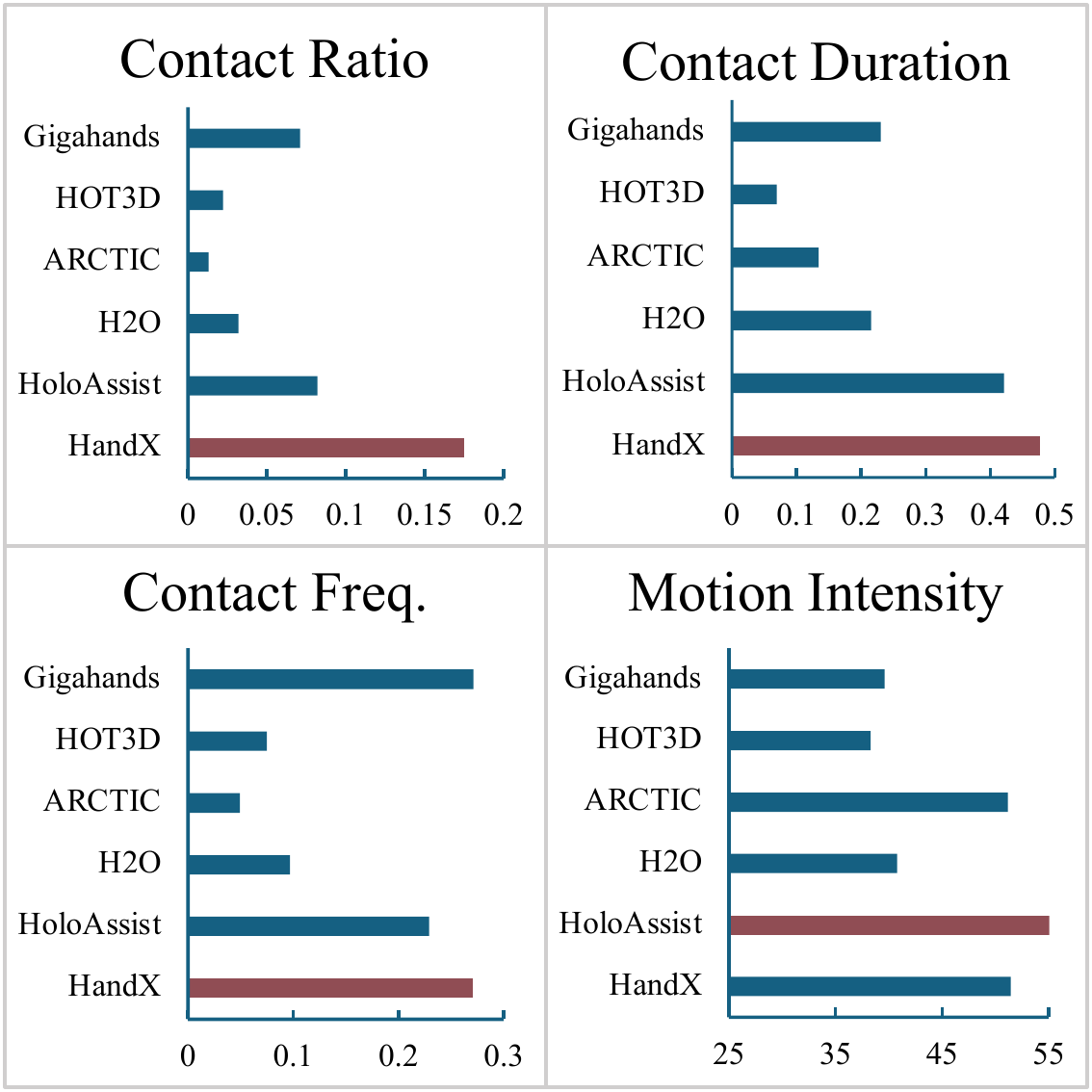}
    \end{minipage}

    \caption{
    \textbf{Comparison of major hand motion datasets.}
    \textbf{Left:} Dataset scale. Values are reported as \textit{HQ (Raw)} where “HQ” denotes high-quality filtered data (Sec.~\ref{sec:Intensity Aware}) and “Raw” (when available) indicates the original data. ``Coarse'' denotes short descriptions without articulation detail, while ``action'' denotes only categorical labels. \oursx provides fine-grained, multi-level language descriptions.
    \textbf{Right:} Statistics of bimanual motion quality. Metrics are defined in Sec.~\ref{sec:dataset_metrics}. \oursx provides contact-rich bimanual motions.
    }
    \label{tab:dataset_comparison_new}
    \vspace{-1em}
\end{table*}

\noindent\textbf{Capturing New Data.}
We collect new data using a 36-camera OptiTrack optical motion-capture system in a dedicated studio, which provides dense coverage for complex bimanual interactions with occlusion and rapid finger motion. Each actor wears 25 reflective hand markers to capture fine-grained articulation of the wrist, palm, fingers, and fingertips (Figure~\ref{fig:mocap_setup}). From the resulting marker trajectories, we reconstruct the hand skeleton by estimating joint centers and enforcing \textit{anatomical constraints} on bone lengths, with \textit{per-frame refinement} for improved kinematic consistency. Additional details on the \textit{studio setup} and \textit{optimization} are provided in Sec.~\ref{sec:mocap_supp} of the supplementary material.

%% file: sec/4_auto_annotation_v5.tex
\section{Bimanual Motion Captioning}\label{sec:3_auto_annotation}

Given the scale of our dataset (Table~\ref{tab:dataset_comparison_new}), manually annotating bimanual motion sequences is prohibitively expensive. Many large foundation models are strong at language understanding and generation; they are inherently text-centric and are not directly effective in modeling continuous, high-dimensional motion data.
To address this challenge, we propose an automatic annotation framework with two stages: (\textbf{a}) extract structured kinematic features from raw hand motion motivated by~\cite{delmas2022posescript,yazdian2023motionscript}, and (\textbf{b}) use a large language model (LLM) to reason over these features and generate coherent textual descriptions. As summarized in Table~\ref{tab:dataset_comparison_new} and illustrated in Figure~\ref{fig:teaser}, this framework enables \oursx to produce large-scale, multi-level, and fine-grained annotations. Unlike template-based labeling~\cite{cha2024text2hoi}, our method generates descriptions grounded in motion dynamics while introducing diversity. Compared with concurrent work~\cite{thambiraja2026clutch,bensabath2025text}, our annotations further capture fine-grained bimanual interactions, especially detailed hand-hand relations.

\noindent\textbf{Kinematic Feature Extraction.}
The goal of kinematic feature extraction is to convert high-dimensional, continuous bimanual motion sequences into structured, semantically meaningful representations that LLMs can reliably interpret. (\textbf{a}) We first compute a set of kinematic \textit{descriptors}, \eg, finger flexion and finger-palm distances, which characterize the detailed pose of both hands \textit{at each frame}, along with their inter-hand spatial relationships, in a structured form. (\textbf{b}) We then analyze the temporal evolution of these descriptors by segmenting the motion into \textit{events}, where each event corresponds either to a change or to a stable interval of a descriptor. This event-based representation captures both dynamic transitions and steady states over time. We organize the events into a structured JSON format (Figure~\ref{fig:feature_json_example}), making them readily accessible for LLM parsing and interpretation. Formal definitions of the descriptors and details of descriptor computation and event extraction are provided in Sec.~\ref{sec:rule-based feature extraction} of the supplementary material.

\noindent\textbf{Translating Kinematic Features into Natural Language.}
Building on the structured kinematic features described above, we leverage the semantic reasoning and generation capabilities of LLMs to produce diverse textual annotations for each motion sequence. Specifically, given the JSON-formatted kinematic features, we design a prompt, shown in Figure~\ref{fig:prompt}, to guide the LLM in generating detailed motion descriptions. The prompt is built around three key principles: (\textbf{a}) explicitly describing the \textit{left hand}, \textit{right hand}, and their \textit{inter-hand relationships} to ensure complete coverage of both local articulations and global coordination patterns; (\textbf{b}) requiring the model to report critical motion events such as contact, separation, and hyperextension; and (\textbf{c}) incorporating temporal context to preserve the sequential progression of motion events.
To increase annotation diversity, we instruct the LLM to generate five levels of textual descriptions with progressively richer detail. These include (\textbf{a}) concise summaries that focus on the most salient movements; (\textbf{b}) balanced descriptions with moderate detail; and (\textbf{c}) comprehensive descriptions that cover all major events, including subtle changes and motion speed variations.

%% file: sec/5_method_v2.tex
\section{Bimanual Motion Generation}

\noindent\textbf{Problem Formulation.}
We denote a two-hand motion sequence with $F$ frames as $\boldsymbol{p} = \{\boldsymbol{p}^1, \boldsymbol{p}^2, \dots, \boldsymbol{p}^F\}$, where $\boldsymbol{p}^i \in \mathbb{R}^{2J \times 3}$ represents the 3D coordinates of all joints from both hands at frame $i$, and $J$ is the number of joints per hand. As detailed in Sec.~\ref{sec:3_auto_annotation}, text prompts can be defined as $T = \{T_L, T_R, T_I\}$, where $T_L$, $T_R$, and $T_I$ describe the left-hand, right-hand, and inter-hand motion, respectively. Our goal is to generate a two-hand motion sequence that is consistent with the text descriptions $T$. For visualization, we optionally recover the MANO parameters~\cite{MANO:SIGGRAPHASIA:2017} through post-optimization to obtain the hand meshes. In the following, we benchmark two representative classes of generative models: diffusion models and autoregressive models.

\subsection{Diffusion Model}
\label{method:diffusion model}

\noindent\textbf{Additional Rotation Scalar in Motion Representation.} We represent each hand joint using both its 3D coordinates and a compact rotation scalar. Given that hand joints have limited rotational degrees of freedom, a single scalar is sufficient. The computation is detailed in Sec.~\ref{sec:diff_data} of the supplementary material. At each frame $i$, we concatenate the joint coordinates and rotation scalars:
\(
\boldsymbol{x}^i = [\boldsymbol{p}^i; \boldsymbol{s}^i] \in \mathbb{R}^{2J \times 4},
\)
yielding a sequence representation $\boldsymbol{x} \in \mathbb{R}^{F \times 2J \times 4}$, where $\boldsymbol{s}^i$ denotes the corresponding 1-DoF rotation scalars.

\begin{figure}[t]
    \centering
    \includegraphics[width=\linewidth]{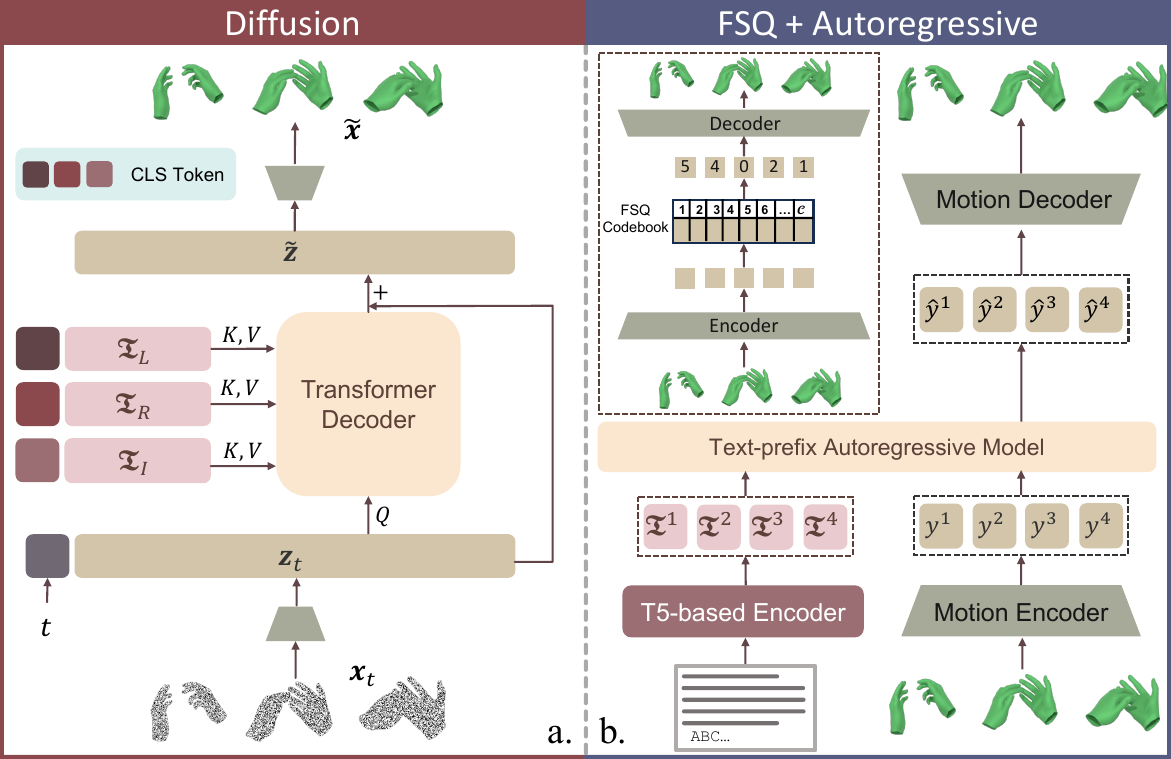}
    \caption{\textbf{Two benchmark models.} (\textbf{a}) Diffusion model. Text embeddings for the left hand, right hand, and bimanual interaction are separately cross-attended with noisy motion embeddings, and then fused through residual connections to predict denoised motion embeddings. (\textbf{b}) Autoregressive model, consisting of Finite Scalar Quantization (FSQ) and a text-prefix autoregressive model. Unlike the diffusion model, it concatenates the left-hand, right-hand, and bimanual text descriptions with separator tokens to form a text prefix, and formulates bimanual motion generation as a token prediction task over motion tokenized by FSQ.}
    \label{fig:pipeline} 
    \vspace{-1em}
\end{figure}

\noindent \textbf{Model Architecture.}
Our diffusion model is trained to iteratively denoise motion sequences. Following \cite{xu2023interdiff}, we train a neural network $\mathcal{G}$ to directly predict the clean signal $\tilde{\boldsymbol{x}}$ from its noisy version $\boldsymbol{x}_t$ at timestep $t$. The noisy input $\boldsymbol{x}_t$ is obtained from the clean motion $\boldsymbol{x}_0$ through the forward diffusion process~\cite{ho2020denoising}:
\(
\boldsymbol{x}_t = \sqrt{\bar{\alpha}_t}\boldsymbol{x}_0 + \sqrt{1 - \bar{\alpha}_t}\boldsymbol{\epsilon},
\)
where $\bar{\alpha}_t = \prod_{t'=1}^t (1-\beta_{t'})$, $\beta_{t'}$ denotes the noise variance, and $\boldsymbol{\epsilon}\sim \mathcal{N}(\boldsymbol{0}, \boldsymbol{I})$.
Given the noisy motion $\boldsymbol{x}_t$ at denoising timestep $t$ and the text prompts $T = (T_L, T_R, T_I)$, the network $\mathcal{G}(\boldsymbol{x}_t, t, T)$ predicts the clean signal $\tilde{\boldsymbol{x}}$.

As illustrated in Figure~\ref{fig:pipeline}(\textbf{a}), we first use an MLP-based encoder $F$ to project the motion representation at each frame into a $D$-dimensional embedding:
\(
\boldsymbol{z}_t = F(\boldsymbol{x}_t) \in \mathbb{R}^{F \times D}.
\)
Following~\cite{tevet2022human}, we further encode the timestep using an MLP-based timestep encoder to obtain a timestep token $\boldsymbol{\mathfrak{t}}$, which is concatenated with the motion embeddings:
\(
\boldsymbol{z}_t' = [\boldsymbol{\mathfrak{t}}; \boldsymbol{z}_t] \in \mathbb{R}^{(1+F)\times D}.
\)
We adopt T5~\cite{raffel2020exploring} as the sequence-to-sequence text encoder for the prompts. We observe that simply concatenating the three types of prompts degrades performance, \eg, the generated motion may assign right-hand movements to the left hand. To address this issue, we encode three types of prompts separately and add a learnable CLS token to each, allowing the model to distinguish left-hand, right-hand, and inter-hand interactions. The resulting three text embeddings are then cross-attended with $\boldsymbol{z}_t'$ and fused through residual connections:
\(
\tilde{\boldsymbol{z}} = \boldsymbol{z}_t' + \sum_{k\in \{L, R, I\}} \mathrm{CrossAttention}(\boldsymbol{z}_t', \boldsymbol{\mathfrak{T}}_k),
\)
where $\boldsymbol{\mathfrak{T}}_k$ ($k \in \{L, R, I\}$) denotes the text embedding for each prompt.
Finally, an MLP-based decoder $G$ maps the fused representation back to motion:
\(
\tilde{\boldsymbol{x}} = G(\tilde{\boldsymbol{z}}) \in \mathbb{R}^{F \times 2J \times 4}.
\)

\begin{figure*}[t]
  \centering

  \includegraphics[width=\textwidth]{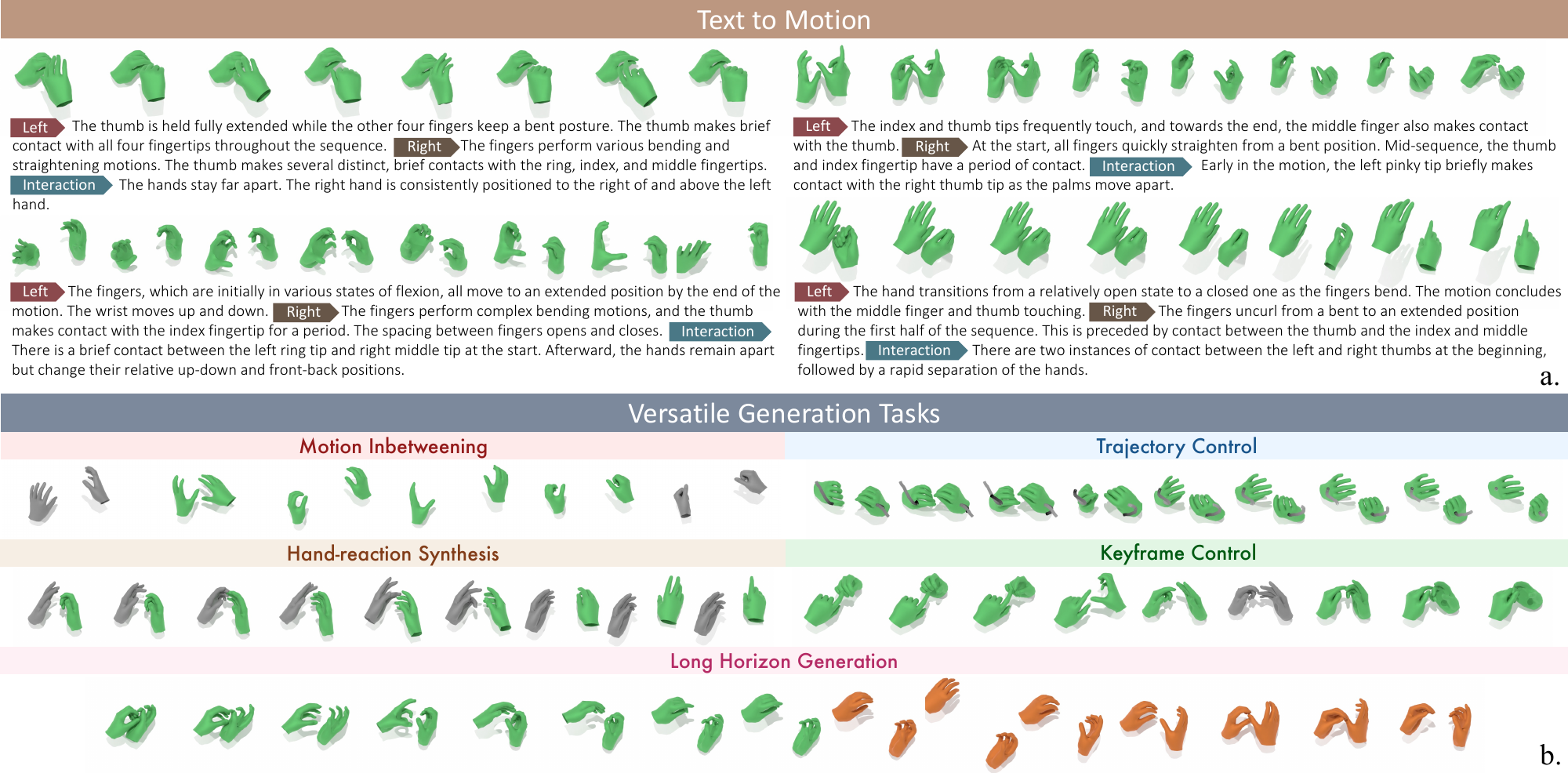}
  
  \caption{\textbf{Qualitative results} of our unified framework, showing (\textbf{a}) high-fidelity text-to-motion generation with fine-grained articulation and contact, and (\textbf{b}) bimanual motion synthesis given versatile spatiotemporal conditions. Gray hands denote the input condition, green hands denote the generation, and orange hands denote the extended long-horizon generation.}
  \label{fig:t2m}
  \vspace{-1em}
\end{figure*}

\noindent \textbf{Versatile Bimanual Motion Generation.} Our framework's design enables a suite of versatile generation tasks from a {single model}. This versatility stems from an inference-time {partial denoising} strategy, which enforces known constraints by blending the input condition with the current sample $\boldsymbol{x}_t$ at each denoising step.
As shown in Figure~\ref{fig:t2m}, our mechanism can achieve comprehensive spatiotemporal and conditional control, such as fixing start and end poses for \textit{Motion In-betweening}, fixing sparse keyframes for \textit{Keyframe-based Generation}, fixing wrist paths for \textit{Wrist Trajectories Generation}, and fixing one hand for \textit{Hand-reaction Synthesis}. The mechanism can also achieve \textit{Long Horizon Generation} by applying partial denoising autoregressively. Implementation details are provided in Sec.~\ref{sec:vers} of the supplementary material.

\subsection{Autoregressive Model}

\noindent\textbf{Overview.}
Autoregressive (AR) modeling is another classic approach, which we benchmark, as illustrated in Figure~\ref{fig:pipeline}(\textbf{b}). Since AR modeling requires discrete motion tokens, we adopt Finite Scalar Quantization (FSQ) as it offers better codebook utilization, reconstruction quality, and scaling behavior~\cite{lu2025scamo}. In the following, we first introduce the motion representation, and then describe the architectures of the motion tokenizer and the autoregressive model.

\noindent \textbf{Motion Representation.}
Unlike the global representation used in the diffusion model (Sec.~\ref{method:diffusion model}), we adopt a local motion representation to \textit{improve codebook utilization}. Specifically, we define the representation at frame $i$ as
\(
    \boldsymbol{x}^i =
    [ \boldsymbol{d}_r^i; \boldsymbol{v}_r^i; \boldsymbol{\theta}_r^i;
    \boldsymbol{p}_l^i; \boldsymbol{v}_l^i; \boldsymbol{s}^i ].
\)
Here, $\boldsymbol{d}_r^i \in \mathbb{R}^{3}$ denotes the relative vector from the left wrist to the right wrist, and $\boldsymbol{v}_r^i \in \mathbb{R}^{3}$ denotes the linear velocity of the right wrist. $\boldsymbol{\theta}_r^i \in \mathbb{R}^{2 \times 6}$ represents the orientations of both wrists. $\boldsymbol{p}_l^i \in \mathbb{R}^{2 \times (J-1) \times 3}$ denotes the local joint positions of both hands with respect to their wrist joints, while $\boldsymbol{v}_l^i \in \mathbb{R}^{2 \times (J-1) \times 3}$ denotes the corresponding local joint velocities. Finally, $\boldsymbol{s}^i \in \mathbb{R}^{2 \times (J-1)}$ denotes the rotation scalars defined in Sec.~\ref{method:diffusion model}.

\noindent \textbf{Motion Tokenizer.} 
Our motion tokenizer consists of a motion encoder $\mathcal{E}$, a motion decoder $\mathcal{D}$, and a finite scalar quantizer $\mathcal{Q}$, following~\cite{lu2025scamo,fan2025go}. The input motion $\boldsymbol{x} = \{\boldsymbol{x}^{i}\}_{i=1}^{F}$ is first encoded by the encoder $\mathcal{E}$ to produce the latent feature $\boldsymbol{y} = \{\boldsymbol{y}^{i}\}_{i=1}^{\lfloor F/l\rfloor}$, where $l$ is the downsample factor. Subsequently, the latent is discretized into $L$ uniformly spaced integer levels as:
\(
    \hat{\boldsymbol{y}} = \mathcal{Q}(\boldsymbol{y}) = 
    \mathrm{round}\!\left({\color{teal}\sigma}(\boldsymbol{y}) \cdot (L - 1)\right),
\)
where $\sigma$ is the sigmoid function and \(L\) defines the number of quantization levels. The optimization objective is defined as
\(\mathcal{L} = \|\boldsymbol{x} - \mathcal{D}(\boldsymbol{\hat{\boldsymbol{y}}})\|_2^2\).

\noindent \textbf{Autoregressive Modeling.} We adopt a text-prefix autoregressive model. As illustrated in Figure~\ref{fig:pipeline}(\textbf{b}), given the text prompts $T = (T_L, T_R, T_I)$, we apply positional encoding and feed them into T5-based encoder to obtain text-prefix latent tokens $\boldsymbol{\mathfrak{T}} = \{\boldsymbol{\mathfrak{T}}^{k}\}_{k=1}^{n_{t}}$, where $n_t$ denotes the number of text tokens.
Motion generation is then formulated as autoregressive next-token prediction, where the model predicts the next motion token $\hat{\boldsymbol{y}}^{k}$ conditioned on the preceding motion latents ${\boldsymbol{y}}^{<k}$ and the text prefix $\boldsymbol{\mathfrak{T}}$.
Following~\cite{lu2025scamo,fan2025go}, attention among text prefix tokens is bidirectional, while attention in the motion branch is causal. The text-prefix autoregressive model is trained with:
\(
\mathcal{L} = - \sum_{k=1}^{n} \log p(\hat{\boldsymbol{y}}^k \mid \boldsymbol{y}^{<k}, \boldsymbol{\mathfrak{T}}),
\)
where $n$ denotes the number of motion tokens.

%% file: sec/experiments.tex
\section{Experiments} \label{sec:experiments}

\subsection{Implementation Details}
To study scaling behavior with respect to both data volume and model capacity, we conduct experiments across multiple training-set sizes and model configurations. For data scaling, we use 5\%, 20\%, and 100\% of the full training set, where the 5\% and 20\% subsets are obtained by uniform random sampling. All models are evaluated on the same validation split for a fair comparison. All model configurations are summarized in Table~\ref{tab:configuration_of_diffusion_models_AR}.
For the diffusion model, we evaluate four model sizes with 4, 8, 12, and 16 Transformer decoder layers. For the autoregressive (AR) model, the tokenizer uses 1D convolutional blocks in both the encoder and decoder, with a temporal downsampling factor of 2. Within this framework, we study multiple model configurations by varying the number of Transformer layers (8, 12, and 16) and the codebook size (512, 1,024, 2,048, and 4,096). During inference, we use deterministic decoding, selecting the token with the highest predicted probability at each step.

\subsection{Metrics} \label{sec:metrics}
Following~\cite{guo2022generating}, we evaluate the realism and diversity of generated hand motion, their alignment with textual descriptions.
To assess realism and diversity, we employ the Fréchet Inception Distance (\textbf{FID}), which quantifies the similarity between the feature distributions of generated and ground truth sequences, and the \textbf{Diversity} metric that measures variability across generated hand motion.  
For textual alignment, we adopt \textbf{R-Precision} and the Multimodal Distance (\textbf{MM Dist}), which quantify feature-level correspondence between generated hand motion and their associated text embeddings.

For bimanual motion generation, traditional metrics are insufficient to assess the quality of hand contact and interaction. We therefore adopt contact precision ($C_\mathrm{prec}$), recall ($C_\mathrm{rec}$), and F1 score ($C_\mathrm{F1}$) to evaluate hand contact accuracy. Contact labels are extracted directly from the ground truth interaction annotations. Specifically, when a contact event occurs in the ground truth, we expect the generated sequence to reproduce the same event at the corresponding frames; successful matches are counted as positive. We empirically set the contact threshold to 2 cm. Additional details are provided in Sec.~\ref{sec:eval_details} of the supplementary material.

\input{table/ablation_diffusion_size}

\begin{table*}[!t]
    \centering
    \caption{\textbf{Ablation study} on the codebook size of FSQ and the model size of autoregressive models. For R-precision, we adopt a batch size of 32. Both the FSQ and autoregressive models are trained on the full training dataset. The primary metrics, \eg, FID, achieve the best performance when both model capacity and codebook size are scaled up. In contrast, scaling only one while keeping the other fixed can degrade performance. For example, R-precision is highest when the autoregressive model size is comparable to the codebook size.}
    \label{tab:ablation_VQ_AR}
    \resizebox{0.85\textwidth}{!}{%
    \begin{tabular}{@{}ccccccccccc@{}}
        \toprule
        \multirow{2}{*}{Model Size(M)} &
        \multirow{2}{*}{Codebook Size} &
        \multicolumn{3}{c}{R-Precision$^\uparrow$} &
        \multirow{2}{*}{FID$^\downarrow$} &
        \multirow{2}{*}{Diversity$^\rightarrow$} &
        \multirow{2}{*}{Matching Dist$^\downarrow$} &
        \multicolumn{3}{c}{Intra-hand Interaction$^\uparrow$} \\
        \cmidrule(lr){3-5} \cmidrule(lr){9-11}
        & & Top 1 & Top 2 & Top 3 & & & &
        $C_\mathrm{prec}$ & $C_\mathrm{rec}$ & $C_\mathrm{F1}$ \\
        \midrule
        \multicolumn{2}{c}{Ground Truth} &
        0.854 & 0.925 & 0.948 & 0.000 & 6.887 & 4.360 & 0.984 & 0.984 & 0.984 \\
        \midrule
        4.63 & 512 & 0.366 & 0.495 & 0.569 & 8.377 & 5.504 & 5.440 & 0.935 & 0.357 & 0.514 \\ 
        26.33 & 512 & 0.277 & 0.401 & 0.495 & 5.071 & 5.872 & 5.556 & 0.850 & 0.422 & 0.561 \\ 
        29.63 & 512 & 0.210 & 0.327 & 0.402 & 4.683 & 6.031 & 5.828 & 0.795 & 0.419 & 0.545 \\
        38.95 & 512 & 0.285 & 0.398 & 0.480 & 5.131 & 5.985 & 5.591 & 0.841 & 0.408 & 0.546 \\
        \midrule
        4.63 & 1,024 & \textbf{0.384} & \textbf{0.518} & \textbf{0.593} & 5.916 & 5.622 & \textbf{5.365} & 0.871 & 0.417 & 0.561 \\
        26.33 & 1,024 & 0.322 & 0.458 & 0.547 & 2.750 & 6.113 & 5.414 & 0.778 & \textbf{0.523} & 0.624 \\
        29.63 & 1,024 & 0.236 & 0.361 & 0.438 & 3.459 & 6.132 & 5.764 & 0.810 & 0.402 & 0.534 \\
        38.95 & 1,024 & 0.328 & 0.461 & 0.536 & 2.812 & 6.155 & 5.442 & 0.793 & 0.521 & \textbf{0.627} \\
        \midrule
        4.63 & 2,048 & 0.329 & 0.458 & 0.536 & 9.138 & 4.988 & 5.471 & 0.845 & 0.361 & 0.504 \\
        26.33 & 2,048 & 0.252 & 0.364 & 0.444 & 3.188 & 6.052 & 5.603 & 0.748 & 0.488 & 0.589 \\
        38.95 & 2,048 & 0.305 & 0.435 & 0.522 & 3.245 & 6.146 & 5.472 & 0.785 & 0.498 & 0.607 \\
        92.27 & 2,048 & 0.182 & 0.288 & 0.354 & 2.949 & 6.156 & 5.882 & 0.694 & 0.493 & 0.574 \\
        \midrule
        4.63 & 4,096 & 0.312 & 0.423 & 0.502 & 9.934 & 5.005 & 5.639 & \textbf{0.947} & 0.216 & 0.347 \\
        26.33 & 4,096 & 0.281 & 0.401 & 0.492 & 3.023 & 5.915 & 5.519 & 0.848 & 0.428 & 0.566 \\
        38.95 & 4,096 & 0.205 & 0.313 & 0.392 & 2.637 & 6.048 & 5.662 & 0.811 & 0.451 & 0.577 \\
        92.27 & 4,096 & 0.134 & 0.221 & 0.283 & 3.050 & 6.025 & 5.953 & 0.754 & 0.376 & 0.497 \\
        215.31 & 4,096 & 0.281 & 0.397 & 0.481 & \textbf{1.721} & \textbf{6.335} & 5.667 & 0.785 & 0.497 & 0.605 \\
        \bottomrule
    \end{tabular}}
    \vspace{-1.2em}
\end{table*}

\input{table/ablation_model_size}

\begin{figure}[!t]
    \centering
    \includegraphics[width=\linewidth]{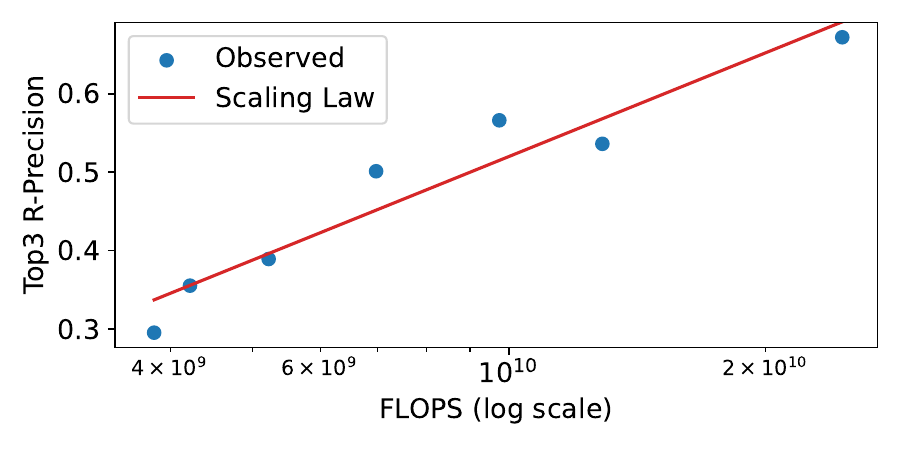}
    \vspace{-1.5em}
    \caption{\textbf{Scaling trend of computational scale.} We observe a clear log-linear relationship between R-precision and FLOPS, with a high correlation coefficient of $0.96$. R-Precision is evaluated with a batch size of 16.}
    \label{fig:fitting scaling equation}
    \vspace{-1.2em}
\end{figure}
\subsection{Quantitative Evaluation}
We analyze how training data scale and model capacity affect performance. Overall, both diffusion and autoregressive models show clear \textit{positive scaling trends}: increasing data and capacity generally improves text-motion alignment and hand-contact quality, although the gains are not strictly monotonic for every metric.

For diffusion models (Table~\ref{tab:ablation_model_size}), scaling either model depth or training data consistently improves the primary metrics, especially R-Precision and contact-related scores. This indicates that better text conditioning and stronger bimanual interaction modeling benefit from both additional capacity and additional supervision. The 12-layer model achieves the best overall contact performance, suggesting that moderate scaling is particularly effective. However, \textit{scaling is not unbounded}. When we further increase the model size to an ultra-large variant with $6.7\times$ more parameters than the 12-layer model, performance drops across all metrics, indicating a clear saturation point beyond which extra capacity no longer helps.

For the autoregressive model (Table~\ref{tab:ablation_VQ_AR}), we find that increasing the FSQ codebook size alone does not reliably improve performance, whereas \textit{jointly} increasing codebook size and model size yields the strongest results. This suggests that finer discrete representations are only beneficial when matched with sufficient autoregressive capacity.

To better characterize the scaling trend, we run a denser set of diffusion model experiments under a fixed 5\% data budget (Table~\ref{tab:denser_configuration}). As shown in Figure~\ref{fig:fitting scaling equation}, Top-3 R-Precision follows an approximately log-linear relationship with FLOPs:
\(
    \mathrm{Rprec} = 0.4391 \times \log_{10}(\mathrm{FLOPS}) - 3.8707.
\)

Overall, these results show that our benchmark supports meaningful scaling in both data and model sizes, but only within an appropriate regime: matched increases in data and capacity improve motion quality, text alignment, and contact coherence, while over-scaling the model alone leads to diminishing or negative returns.

\begin{figure}[!t]
    \centering
    \includegraphics[width=\linewidth]{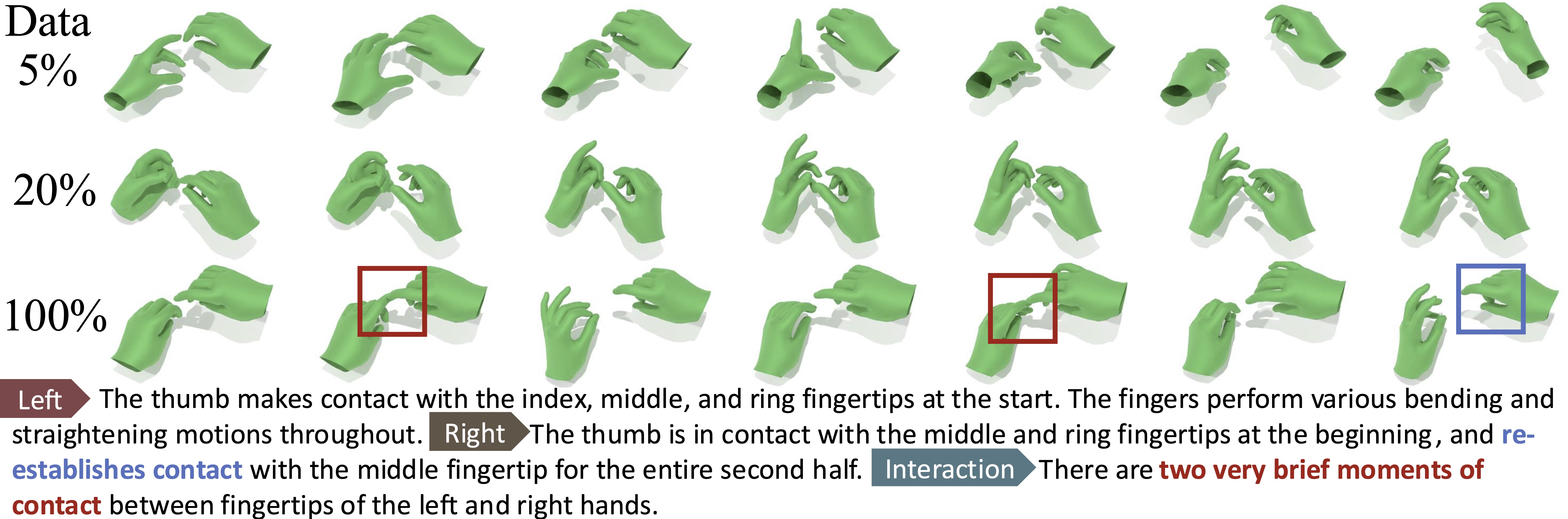}
    \includegraphics[width=\linewidth]{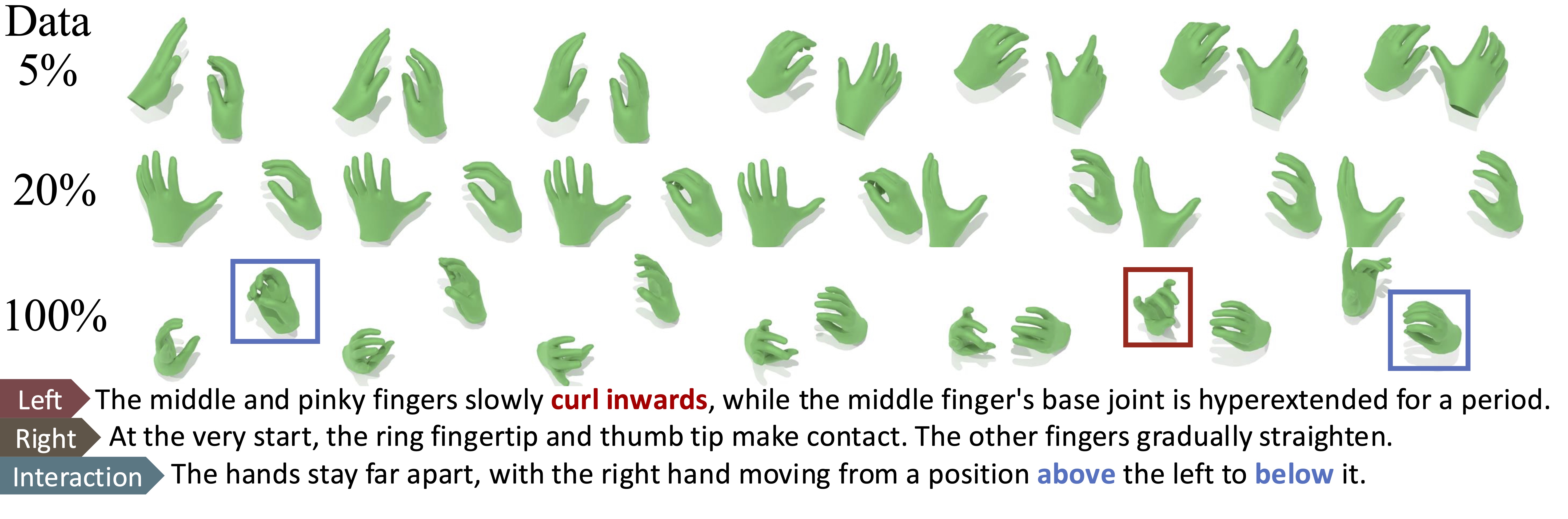}
\caption{Qualitative comparison of diffusion models trained with different \textbf{data scales}. The model trained on the full dataset generates more expressive motion with better text alignment.}
    \label{fig:data_scaling_1}
\end{figure}
\begin{figure}[!t]
    \centering
    \includegraphics[width=\linewidth]{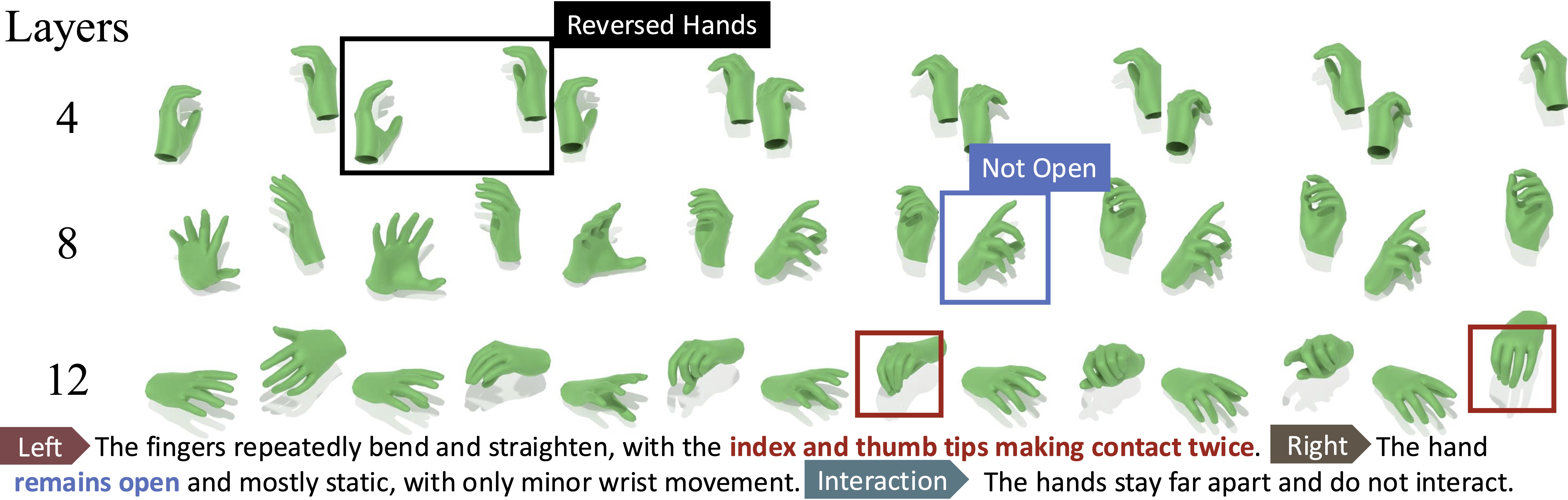}
    \includegraphics[width=\linewidth]{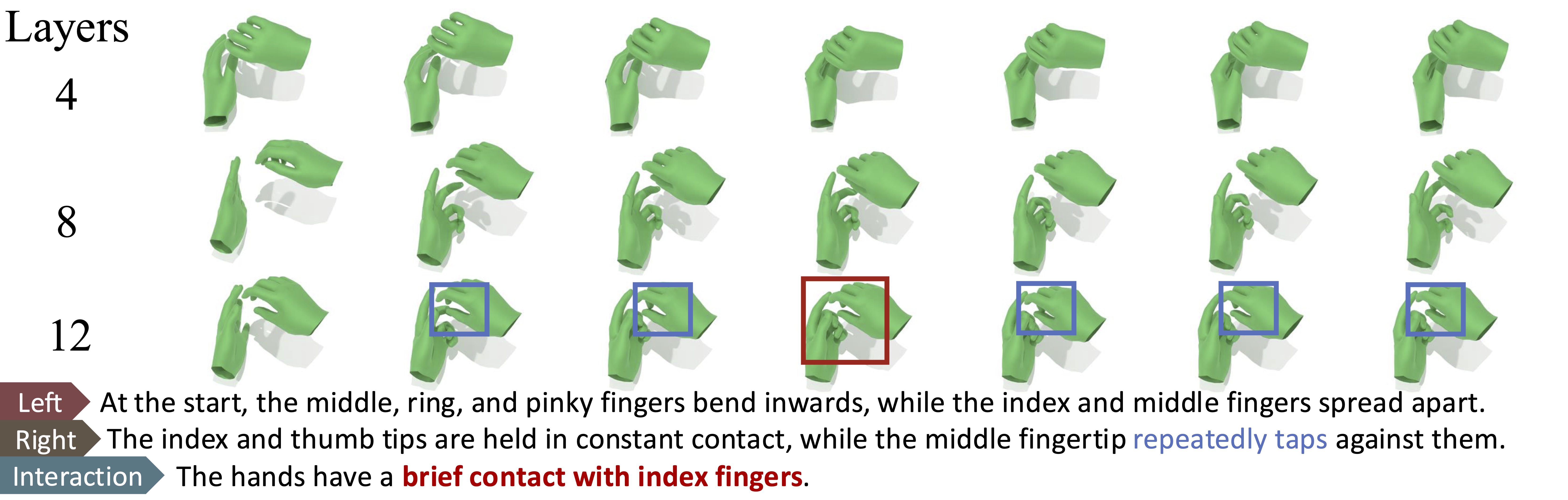}
\caption{Qualitative comparison of diffusion models with different \textbf{model scales}. Larger models produce motion that is better aligned with the text and exhibits improved bimanual contact.}
    \label{fig:model_scaling_1}
    \vspace{-1.2em}
\end{figure}
\subsection{Qualitative Evaluation}
We provide qualitative visualizations of the performance of our model in Figure~\ref{fig:t2m}. The visualizations highlight our model's ability to synthesize fine-grained finger articulation and realistic inter-hand coordination, successfully capturing complex contact events specified in the text prompt, while supporting a wide range of generation tasks. 

We further provide qualitative comparisons to illustrate these scaling trends in diffusion models. Figure~\ref{fig:data_scaling_1} compares samples generated by models trained with different amounts of data, highlighting the visual impact of the dataset scale. Figure~\ref{fig:model_scaling_1} compares samples from models with different numbers of decoder layers, showing how increased model capacity affects generation fidelity. Key visual differences are highlighted with colored boxes, and the corresponding textual cues are marked in bold using the same color.

%% file: table/ablation_diffusion_size.tex
\begin{table*}[!t]
    \centering
    \caption{\textbf{Ablation study} on model size and data size. For R-precision, we adopt a batch size of 32. We observe clear scaling trends for our primary metrics, \eg, {R-Precision improves consistently} as we scale both data and model sizes, while {Intra-hand $C_\mathrm{F1}$} shows a strong positive trend, culminating in the best performance with 12 layers of decoder and all training data.}
    \label{tab:ablation_model_size}
    \resizebox{\textwidth}{!}{%
    \begin{tabular}{@{}ccccccccccc@{}}
        \toprule
        \multirow{2}{*}{Dataset Ratio} & 
        \multirow{2}{*}{Decoder Layers} & 
        \multicolumn{3}{c}{R-Precision$^\uparrow$} & 
        \multirow{2}{*}{FID$^\downarrow$} & 
        \multirow{2}{*}{Diversity$^\rightarrow$} & 
        \multirow{2}{*}{Matching Dist$^\downarrow$} & 
        \multicolumn{3}{c}{Intra-hand Interaction$^\uparrow$} \\
        \cmidrule(lr){3-5} \cmidrule(lr){9-11}
        & & Top 1 & Top 2 & Top 3 & & & & 
        $C_\mathrm{prec}$ & $C_\mathrm{rec}$ & $C_\mathrm{F1}$ \\
        \midrule

        \multicolumn{2}{c}{Ground Truth} & 
        $0.854^{\pm0.094}$ & $0.925^{\pm0.059}$& $0.948^{\pm0.043}$ & 
        $0.000^{\pm0.000}$ & 
        $6.887^{\pm0.078}$ & 
        $4.360^{\pm0.373}$ & 
        $0.984^{\pm0.000}$ & 
        $0.984^{\pm0.000}$ & 
        $0.984^{\pm0.000}$ \\
        \midrule

        0.05 & 4 & 
        $0.142^{\pm0.059}$ & $0.228^{\pm0.068}$ & $0.296^{\pm0.071}$ & 
        2.574$^{\pm0.109}$ & 
        7.406$^{\pm0.069}$ & 
        6.581$^{\pm0.158}$ & 
        0.628$^{\pm0.005}$ & 0.447$^{\pm0.006}$ & 0.523$^{\pm0.006}$ \\

        0.05 & 8 & 
        0.223$^{\pm0.071}$ & 0.334$^{\pm0.077}$ & 0.417$^{\pm0.079}$ & 
        1.548$^{\pm0.048}$ & 
        7.161$^{\pm0.054}$ & 
        6.100$^{\pm0.159}$ & 
        0.659$^{\pm0.004}$ & 0.576$^{\pm0.006}$ & 0.615$^{\pm0.005}$ \\

        0.05 & 12 & 
        0.343$^{\pm0.094}$ & 0.485$^{\pm0.092}$ & 0.573$^{\pm0.086}$ & 
        1.837$^{\pm0.033}$ & 
        5.935$^{\pm0.016}$ & 
        ${\pmb{5.331}}^{\pm{\pmb{0.165}}}$\unboldmath & 
        0.701$^{\pm0.002}$ & 0.553$^{\pm0.003}$ & 0.618$^{\pm0.002}$ \\
        \midrule

        0.2 & 4 & 
        0.145$^{\pm0.062}$ & 0.233$^{\pm0.073}$ & 0.301$^{\pm0.080}$ & 
        2.181$^{\pm0.051}$ & 
        6.903$^{\pm0.064}$ & 
        6.412$^{\pm0.238}$ & 
        0.703$^{\pm0.008}$ & 0.379$^{\pm0.008}$ & 0.493$^{\pm0.008}$ \\
        
        0.2 & 8 & 
        0.262$^{\pm0.077}$ & 0.388$^{\pm0.081}$ & 0.466$^{\pm0.084}$ & 
        3.053$^{\pm0.099}$ & 
        7.581$^{\pm0.037}$ & 
        6.242$^{\pm0.164}$ & 
        $\pmb{0.739}^{\pm\pmb{0.003}}$ & 0.460$^{\pm0.009}$ & 0.567$^{\pm0.006}$ \\

        0.2 & 12 & 
        0.357$^{\pm0.083}$ & 0.493$^{\pm0.091}$ & 0.578$^{\pm0.091}$ & 
        \pmb{1.140}$^{\pm\pmb{0.055}}$ & 
        6.770$^{\pm0.026}$ & 
        5.628$^{\pm0.155}$ & 
        0.733$^{\pm0.006}$ & 0.517$^{\pm0.007}$ & 0.606$^{\pm0.004}$ \\
        \midrule

        1.0 & 4 & 
        0.168$^{\pm0.061}$ & 0.259$^{\pm0.071}$ & 0.327$^{\pm0.074}$ & 
        2.219$^{\pm0.126}$ & 
        7.320$^{\pm0.026}$ & 
        6.429$^{\pm0.170}$ & 
        0.704$^{\pm0.007}$ & 0.408$^{\pm0.001}$ & 0.517$^{\pm0.002}$ \\

        1.0 & 8 & 
        0.285$^{\pm0.076}$ & 0.409$^{\pm0.084}$ & 0.491$^{\pm0.085}$ & 
        1.617$^{\pm0.071}$ & 
        7.037$^{\pm0.039}$ & 
        5.924$^{\pm0.148}$ & 
        0.713$^{\pm0.010}$ & 0.476$^{\pm0.006}$ & 0.571$^{\pm0.007}$ \\

        1.0 & 12 & 
        \pmb{0.427}$^{\pm\pmb{0.079}}$ & \pmb{0.554}$^{\pm\pmb{0.076}}$ & \pmb{0.631}$^{\pm\pmb{0.075}}$ & 
        1.349$^{\pm0.014}$ & 
        7.220$^{\pm0.025}$ & 
        5.500$^{\pm0.159}$ & 
        0.693$^{\pm0.005}$ & \pmb{0.596}$^{\pm\pmb{0.007}}$ & $\pmb{0.641}^{\pm\pmb{0.004}}$ \\
        \midrule

        1.0 & 16 & 
        0.382$^{\pm 0.083}$ & 0.519$^{\pm 0.086}$ & 0.603$^{\pm 0.086}$ & 
        1.675$^{\pm 0.024}$ & 
        6.426$^{\pm0.052}$ & 
        5.449$^{\pm0.238}$ & 
        0.722$^{\pm0.003}$ & 0.549$^{\pm0.009}$ & 0.624$^{\pm0.005}$ \\
        \bottomrule
    \end{tabular}}%
\end{table*}

%% file: table/ablation_model_size.tex
\begin{table*}[H]
    \centering
    \caption{\textbf{Ablation study} of number of decoder layers of diffusion model and size of dataset. For R-precision, we adopt a batch size of 32.By comparing models trained on different ratio of our dataset and different decoder layer numbers, we observe clear scaling trends for our primary metrics. {{R-Precision} improves consistently} as we scale both data and model size, while {Intra-hand $C_\mathrm{F1}$} shows a strong positive trend, culminating in the best performance at the largest scale, with 12 layers of decoder and all training data.}
    \label{tab:ablation_model_size}
    \resizebox{\textwidth}{!}{%
    \begin{tabular}{@{}ccccccccccc@{}}
        \toprule
        \multirow{2}{*}{Dataset Ratio} &
        \multirow{2}{*}{Number of Decoder Layers} &
        \multicolumn{3}{c}{R-Precision$^\uparrow$} &
        \multirow{2}{*}{FID$^\downarrow$} &
        \multirow{2}{*}{Diversity$^\rightarrow$} &
        \multirow{2}{*}{Matching Dist$^\downarrow$} &
        \multicolumn{3}{c}{Intra-hand Interaction$^\uparrow$} \\
        \cmidrule(lr){3-5} \cmidrule(lr){9-11}
        & & Top 1 & Top 2 & Top 3 & & & &
        $C_\mathrm{prec}$ & $C_\mathrm{rec}$ & $C_\mathrm{F1}$ \\
        \midrule

        \multicolumn{2}{c}{Ground Truth} &
        0.854$^{\pm0.094}$ & 0.925$^{\pm0.059}$& 0.948$^{\pm0.043}$ & 
        0.000$^{\pm0.000}$ &
        6.887$^{\pm0.078}$ &
        4.360$^{\pm0.373}$ &
        0.984$^{\pm0.000}$ &
        0.984$^{\pm0.000}$ &
        0.984$^{\pm0.000}$ \\
        \midrule

        0.05 & 4 &
        0.142$^{\pm0.059}$ & 0.228$^{\pm0.068}$ & 0.296$^{\pm0.071}$ &
        2.574$^{\pm0.109}$ &
        7.406$^{\pm0.069}$ &
        6.581$^{\pm0.158}$ &
        0.628$^{\pm0.005}$ & 0.447$^{\pm0.006}$ & 0.523$^{\pm0.006}$ \\

        0.05 & 8 &
        0.223$^{\pm0.071}$ & 0.334$^{\pm0.077}$ & 0.417$^{\pm0.079}$ &
        1.548$^{\pm0.048}$ &
        7.161$^{\pm0.054}$ &
        6.100$^{\pm0.159}$ &
        0.659$^{\pm0.004}$ & 0.576$^{\pm0.006}$ & 0.615$^{\pm0.005}$ \\

        0.05 & 12 &
        0.343$^{\pm0.094}$ & 0.485$^{\pm0.092}$ & 0.573$^{\pm0.086}$ &
        1.837$^{\pm0.033}$ &
        5.935$^{\pm0.016}$ &
        \textbf{5.331}$^{\pm\textbf{0.165}}$ &
        0.701$^{\pm0.002}$ & 0.553$^{\pm0.003}$ & 0.618$^{\pm0.002}$ \\

        0.2 & 8 &
        0.262$^{\pm0.077}$ & 0.388$^{\pm0.081}$ & 0.466$^{\pm0.084}$ &
        3.053$^{\pm0.099}$ &
        7.581$^{\pm0.037}$ &
        6.242$^{\pm0.164}$ &
        \textbf{0.739}$^{\pm\textbf{0.003}}$ & 0.460$^{\pm0.009}$ & 0.567$^{\pm0.006}$ \\

        0.2 & 12 &
        0.357$^{\pm0.083}$ & 0.493$^{\pm0.091}$ & 0.578$^{\pm0.091}$ &
        \textbf{1.140}$^{\pm\textbf{0.055}}$ &
        \textbf{6.770$^{\pm0.026}$} &
        5.628$^{\pm0.155}$ &
        0.733$^{\pm0.006}$ & 0.517$^{\pm0.007}$ & 0.606$^{\pm0.004}$ \\

        1.0 & 4 &
        0.168$^{\pm0.061}$ & 0.259$^{\pm0.071}$ & 0.327$^{\pm0.074}$ &
        2.219$^{\pm0.126}$ &
        7.320$^{\pm0.026}$ &
        6.429$^{\pm0.170}$ &
        0.704$^{\pm0.007}$ & 0.408$^{\pm0.001}$ & 0.517$^{\pm0.002}$ \\

        1.0 & 8 &
        0.285$^{\pm0.076}$ & 0.409$^{\pm0.084}$ & 0.491$^{\pm0.085}$ &
        1.617$^{\pm0.071}$ &
        7.037$^{\pm0.039}$ &
        5.924$^{\pm0.148}$ &
        0.713$^{\pm0.010}$ & 0.476$^{\pm0.006}$ & 0.571$^{\pm0.007}$ \\

        1.0 & 12 &
        \textbf{0.427}$^{\pm\textbf{0.079}}$ & \textbf{0.554}$^{\pm\textbf{0.076}}$ & \textbf{0.631}$^{\pm\textbf{0.075}}$ &
        1.349$^{\pm0.014}$ &
        7.220$^{\pm0.025}$ &
        5.500$^{\pm0.159}$ &
        0.693$^{\pm0.005}$ & \textbf{0.596}$^{\pm\textbf{0.007}}$ & \textbf{0.641}$^{\pm0.004}$ \\
        \bottomrule
    \end{tabular}}%
\end{table*}

\begin{table*}[H]
    \centering
    \caption{\textbf{Ablation study} on the codebook size of FSQ and the model size of autoregressive models. For R-precision, we adopt a batch size of 32. Note that following previous works~\cite{lu2025scamo,fan2025go}, we adopt a deterministic
(greedy) decoding strategy during inference, selecting at each step the
token with the highest predicted probability. And the FSQ and autoregressive models are both trained on the full training dataset.}
    \label{tab:ablation_VQ_AR}
    \resizebox{0.8\textwidth}{!}{%
    \begin{tabular}{@{}ccccccccccc@{}}
        \toprule
        \multirow{2}{*}{Model Size(M)} &
        \multirow{2}{*}{Codebook Size} &
        \multicolumn{3}{c}{R-Precision$^\uparrow$} &
        \multirow{2}{*}{FID$^\downarrow$} &
        \multirow{2}{*}{Diversity$^\rightarrow$} &
        \multirow{2}{*}{Matching Dist$^\downarrow$} &
        \multicolumn{3}{c}{Intra-hand Interaction$^\uparrow$} \\
        \cmidrule(lr){3-5} \cmidrule(lr){9-11}
        & & Top 1 & Top 2 & Top 3 & & & &
        $C_\mathrm{prec}$ & $C_\mathrm{rec}$ & $C_\mathrm{F1}$ \\
        \midrule
        \multicolumn{2}{c}{Ground Truth} &
        0.854 & 0.925& 0.948& 0.000
        
         &6.887 
        &4.360
        & 0.984
         & 0.984 & 0.984 \\
        \midrule
        29.63 & 512 & 0.210 & 0.327 & 0.402 & 4.683 & 6.031 & 5.828 & 0.795 & 0.419 & 0.545 \\
         29.63 & 1024 & 0.236 & 0.361 & 0.438 & 3.459 & 6.132 & 5.764 & \textbf{0.810} & 0.402 & 0.534 \\
        92.27 & 2048 & 0.182 & 0.288 & 0.354 & 2.949 & 6.156 & 5.882 & 0.694 & 0.493 & 0.574 \\
        92.27 & 4096 & 0.134 & 0.221 & 0.283 & 3.050 & 6.025 & 5.953 & 0.754 & 0.376 & 0.497 \\
        215.31 & 4096 & \textbf{0.281} & \textbf{0.397} & \textbf{0.481} & \textbf{1.721} & \textbf{6.335}     & \textbf{5.667} & 0.785 & \textbf{0.497} & \textbf{0.605} \\
       
        \bottomrule
    \end{tabular}}%
\end{table*}

%% file: sec/8_conclusion.tex
\section{Conclusion}
In this work, we address the challenge of generating realistic, text-conditioned bimanual hand motion. We introduce \oursx, a large-scale unified dataset built by consolidating diverse motion sources and capturing new high-fidelity, contact-rich bimanual interactions. We further propose an automatic captioning strategy that decouples kinematic feature extraction from LLM-based semantic reasoning, enabling fine-grained, multi-level textual annotations. Based on this dataset, we establish a benchmark with both diffusion and autoregressive models, supporting diverse generation tasks such as motion inbetweening and hand-reaction synthesis. Our experiments reveal clear empirical scaling trends: jointly increasing dataset size and model capacity consistently improves text-motion alignment and contact accuracy. This work provides both a strong dataset foundation and a comprehensive benchmark framework for future research on dexterous hand motion synthesis.

\paragraph{Acknowledgments.} We thank Liuyu Bian for support with the robotic demo implementation. This work was supported in part by Snap Inc., NSF under Grants 2106825 and 2519216, the DARPA Young Faculty Award, the ONR Grant N00014-26-1-2099, and the NIFA Award 2020-67021-32799. This work used computational resources, including the NCSA Delta and DeltaAI and the PTI Jetstream2 supercomputers through allocations CIS230012, CIS230013, CIS240311, and CIS240428 from the Advanced Cyberinfrastructure Coordination Ecosystem: Services \& Support (ACCESS) program, as well as the TACC Frontera supercomputer, Amazon Web Services (AWS), and OpenAI API through the National Artificial Intelligence Research Resource (NAIRR) Pilot. We also thank the Toyota Research Institute for partial support of the robotic hardware used in this research.

%% file: sec/X_suppl.tex
\clearpage
\setcounter{page}{1}
\maketitlesupplementary

\setcounter{table}{0}
\renewcommand{\thetable}{\Alph{table}}
\renewcommand*{\theHtable}{\thetable}
\setcounter{figure}{0}
\renewcommand{\thefigure}{\Alph{figure}}
\renewcommand*{\theHfigure}{\thefigure}
\setcounter{section}{0}
\renewcommand{\thesection}{\Alph{section}}
\renewcommand*{\theHsection}{\thesection}
\setcounter{equation}{0}
\renewcommand{\theequation}{\Alph{equation}}
\renewcommand*{\theHequation}{\theequation}

\noindent This supplementary material is organized as follows. Sec.~\ref{sec:data_const} describes our motion capture system and the construction of \oursx from public data, including quality control procedures. Sec.~\ref{sec:rule-based feature extraction} details our bimanual motion captioning pipeline. Sec.~\ref{sec:diff_data} provides additional details on bimanual motion representations. Sec.~\ref{sec:vers} explains how our model supports versatile motion generation. Finally, Sec.~\ref{sec:eval_details} presents additional details on metrics, evaluation results, and the user study.

\section{\oursx Dataset}
\label{sec:data_const}

\begin{figure}
\includegraphics[width=1\columnwidth]{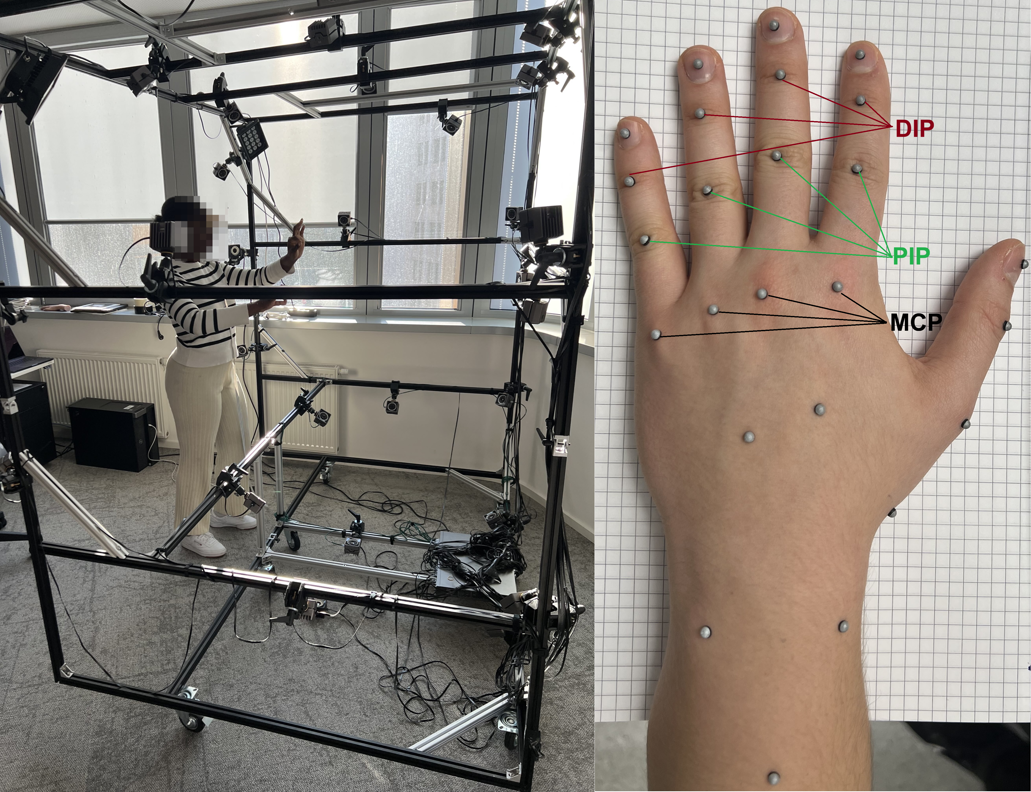}
\caption{Our motion capture configuration: (\textbf{a}) 36-camera OptiTrack studio and (\textbf{b}) placement of 25 markers on each hand. }
\label{fig:mocap_setup}
\end{figure}
\subsection{Motion Capture}\label{sec:mocap_supp}

This subsection details the full data capture pipeline, from motion acquisition to marker-based skeleton reconstruction, and highlights the key design choices that ensure high-quality kinematic data.

\noindent \textbf{Motion Capture Setup.}
We record all motion data using an OptiTrack motion capture system~\cite{optitrack} within a dedicated studio. The capture volume is monitored by 36 high-speed infrared cameras, placed to maximize coverage and minimize marker occlusion during complex two-hand interactions or rapid finger movements, as shown in Figure~\ref{fig:mocap_setup}.

\noindent \textbf{Marker Placement.} Actors have 25 miniature infrared reflective markers (3mm in diameter) glued directly onto the skin of each hand. The placement of these markers (illustrated in Figure~\ref{fig:mocap_setup}(\textbf{b})) is meticulously designed to capture the nuanced articulation of the hand, covering the wrist, the dorsal surface of the palm (metacarpals), three key points on each finger corresponding to the metacarpophalangeal (MCP), proximal interphalangeal (PIP), and distal interphalangeal (DIP) joints, and the fingertip (center of the nail).

\noindent \textbf{Skeleton Reconstruction.} The raw output from the OptiTrack system provides 3D coordinates for the 25 surface markers. To estimate the underlying skeleton joint positions from these surface markers, we first compute the anatomical normal direction $\vec{n}$ for each marker, pointing from the skin surface inward towards the bone. For joints along a finger (\eg, PIP, DIP), this normal is derived from the plane formed by the marker and its proximal and distal neighbors (\eg, using MCP, PIP, and DIP markers). For MCP joints, the normal is computed from the plane defined by neighboring MCP markers on the dorsal surface of the hand.

Next, we estimate the depth $d$ from the skin to the joint center along this normal. This depth value is scaled based on the actor's overall hand size, which is determined during a calibration phase. The final 3D position of a skeleton joint $J_p$ is then calculated by offsetting its corresponding marker position $M_p$ along the computed normal $\vec{n}$ by the estimated depth $d$:
\begin{equation}
 J_p = M_p + \vec{n} \cdot d.\end{equation}

\noindent \textbf{Wrist Optimization.}
While the above process effectively locates the finger joints, the calculations are still affected to soft-tissue artifacts. The most significant non-rigid deformation occurs at the wrist, where skin and soft tissue can stretch and compress substantially, leading to inconsistent distances between, \eg, an MCP marker and a wrist marker.

To compensate for this, we run an iterative optimization process. We assume that the bone lengths, specifically, the distances from each MCP joint to the wrist joint center, remain constant.
First, we calculate a reference ``bone length'' $L_{ref_i}$ for each MCP-to-wrist connection ($i=1,..,5$) from a static, neutral calibration pose.
Then, for every frame in the recording, we iteratively optimize the 3D position of the single wrist joint, $J_{wrist}$, to find the position $J_{wrist}^*$ that minimizes the squared error between the current calculated distances and these reference lengths:
\begin{equation}
     J_{wrist}^* = \operatorname*{argmin}_{J_{wrist}} \sum_{i=1}^{5} \left( \| J_{MCP_i} - J_{wrist} \| - L_{ref_i} \right)^2.
\end{equation}
This optimization ensures that the wrist joint position is anatomically consistent with the relatively rigid palm, effectively filtering out the artifacts caused by soft tissue deformation.

\subsection{Data Consolidation and Unification}
As discussed in Sec.~\ref{sec:dataset}, an important component of \oursx comes from aggregating public datasets, whose licenses are summarized in Sec.~\ref{sec:license}. To ensure consistency across these heterogeneous sources, we unify both the motion representation and the global coordinate system.
First, we map all hand motion to a unified 21-joint skeletal topology, ensuring a consistent joint definition and ordering across datasets. We then canonicalize the global coordinates at the sequence level while preserving a right-handed coordinate frame. After this transformation, in the first frame, the positive $x$-axis points from the left wrist to the right wrist, the positive $y$-axis points from the wrists toward the fingertips, and the positive $z$-axis points upward. This two-stage unification establishes a consistent skeletal and spatial reference across all data sources, facilitating subsequent processing and analysis.

\subsection{Clip Extraction}
During the post-processing stage of \oursx, we divide long motion sequences into clips of 60 frames (2 seconds at 30 FPS). This clip length follows the standard setting in prior work on short-term hand motion modeling~\cite{zhu2023taming,yang2023diffusestylegesture}. It is long enough to capture a complete atomic hand action (\eg, a full pinch-and-release, a grasp, or a sign) while remaining efficient for training.
However, the aggregated data may contain defective frames that either do not provide meaningful bimanual pose or motion information or exhibit abrupt changes across consecutive frames. To ensure data quality, we adopt two principles during clip extraction: \textbf{(a)} detecting and removing defective frames, and \textbf{(b)} using a non-overlapping extraction strategy.

\noindent\textbf{Sequences without defective frames.}
If no defective frames are detected, we segment the sequence into non-overlapping clips using windows $\mathcal{W}(t)$ of length $L$:
\begin{equation}
\begin{aligned}
\mathcal{W}(t) &= \{t,\dots,t+L-1\},\\
t &\in \{0, L, 2L, \dots\},
\end{aligned}
\end{equation}
where $L = 60$ denotes the clip length. This construction ensures that adjacent clips do not overlap.

\noindent\textbf{Sequences with defective frames.}
If defective frames are present, we first remove their indices and then perform segmentation only on the remaining valid continuous intervals. Let $\mathcal{D} \subset \{0,\ldots,T-1\}$ denote the set of defective frames. We partition the valid index set $\{0,\ldots,T-1\}\setminus\mathcal{D}$ into maximal contiguous intervals:
\[
\mathcal{S}=\big\{[a_k,b_k]\big\}_{k=1}^{K_s}, \quad
[a_k,b_k]\cap\mathcal{D}=\varnothing,\;\; b_k-a_k+1 \ge L.
\]
Within each interval $[a_k,b_k]$, we place disjoint windows $\mathcal{W}(t)$ with stride $s=L=60$. This guarantees that every extracted clip is fully valid and that no overlap exists between adjacent clips.

\subsection{Clip Filtering}
\label{sec:Intensity Aware}

As shown in Table~\ref{tab:dataset_comparison_new}(\textbf{b}), analysis of the source datasets reveals a substantial data imbalance in most datasets, particularly in terms of motion intensity and inter-hand interaction, \eg, the hand is idle for most of the time. This imbalance is problematic for training, as it may bias the model toward low-activity states.

\begin{figure}[t]
    \centering
    \includegraphics[width=0.5\textwidth]{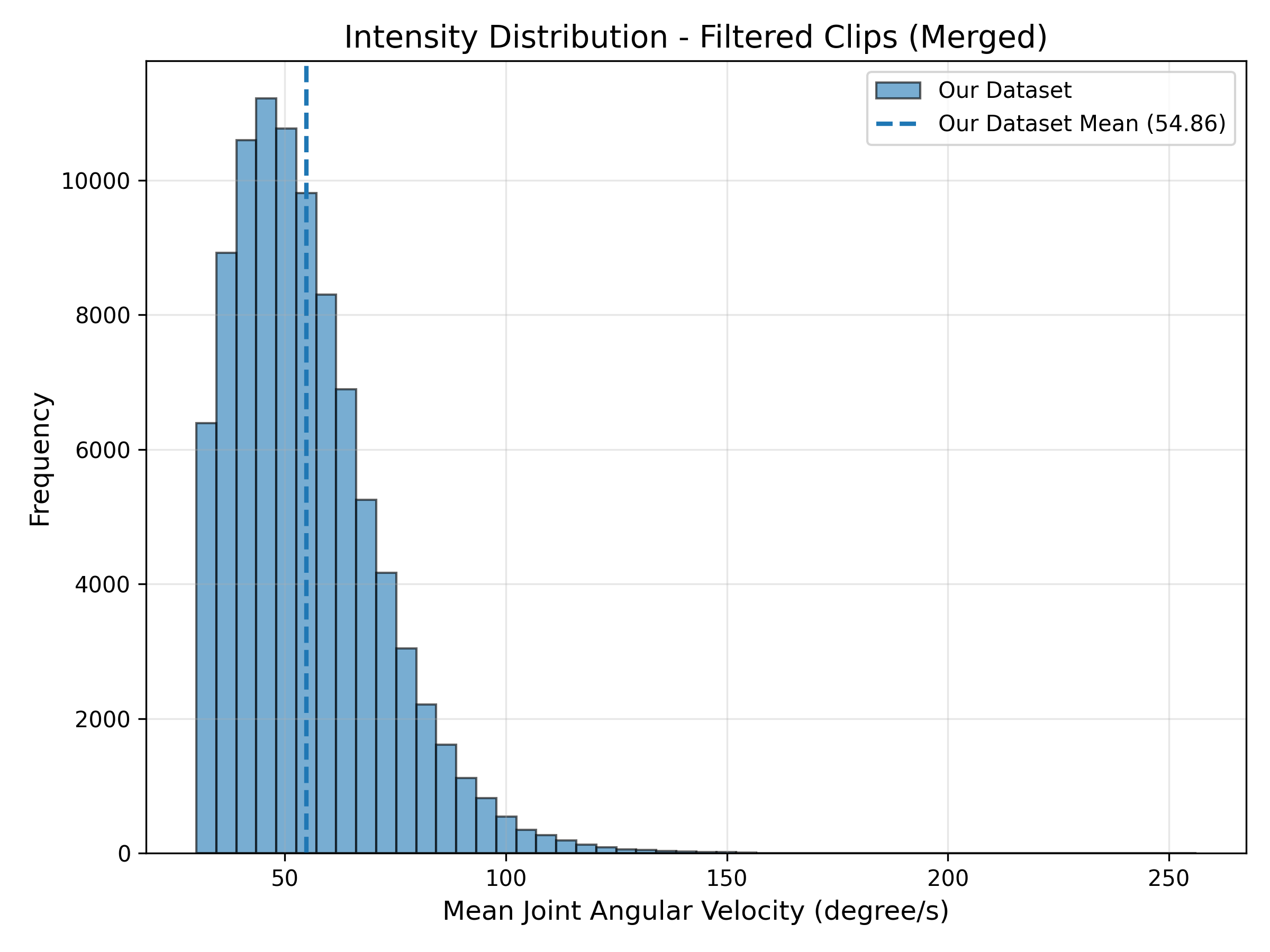}
\caption{\textbf{Distribution of filtered clips.}
    This figure shows the intensity distribution of the filtered clips.}
    \label{fig:filter_distribution}
\end{figure}

To remove static or uninformative segments from the split clips, we apply an \textit{intensity-aware} criterion that identifies and discards static or low-activity segments, retaining only clips with meaningful bimanual dynamics. Specifically, we compute an \textbf{action intensity metric} based on joint angular velocity, which is independent of the global motion of the hands. A clip is retained only if both hands exhibit sufficiently strong motion simultaneously, ensuring that the final dataset consists of highly dynamic bimanual interactions. The resulting distribution of filtered clips is shown in Figure~\ref{fig:filter_distribution}.

\noindent \textbf{Action Intensity Metric.}
Let $\hat{\boldsymbol{v}}_t^{\mathrm{pre}}$ and $\hat{\boldsymbol{v}}_t^{\mathrm{nxt}}$ denote the unit limb directions before and after a joint at time $t$, respectively. The inter-segment angle is computed as
\begin{equation} 
\theta_t=\arccos\big(\hat{\boldsymbol{v}}_t^{\mathrm{pre}}\cdot\hat{\boldsymbol{v}}_t^{\mathrm{nxt}}\big).
\end{equation}

The angular velocity $\omega$ is then computed from the temporal difference of the bending angle for each finger chain.

For each hand $h \in \{\mathrm{left}, \mathrm{right}\}$ with joint set $\mathcal{J}_h$, we assign a hierarchical importance weight $\lambda_j$ to each joint $j \in \mathcal{J}_h$. To emphasize structural pose changes, proximal joints are assigned larger weights than distal ones. The hand-level clip intensity over a temporal window $\mathcal{W}$ is computed as the weighted average of angular velocities:
\begin{equation}
  \bar{\omega}_h(\mathcal{W}) = \frac{\sum_{t\in\mathcal{W}}\sum_{j\in\mathcal{J}_h} \lambda_j \,\omega_t^{(j)}}{\sum_{t\in\mathcal{W}}\sum_{j\in\mathcal{J}_h} \lambda_j}.
\end{equation}

The bimanual clip intensity is then defined as the average intensity of both hands:
\begin{equation}
  \bar{\omega}(\mathcal{W}) = \frac{1}{2}\Big(\bar{\omega}_{\mathrm{left}}(\mathcal{W}) + \bar{\omega}_{\mathrm{right}}(\mathcal{W})\Big).
\end{equation}

\noindent \textbf{Filtering Rule.}
We retain only those clips that exhibit sufficiently strong motion on both hands. Specifically, a clip $\mathcal{W}$ is kept if
\[
\bar{\omega}_{\mathrm{left}}(\mathcal{W})\ge \tau_{\mathrm{hand}},\
\bar{\omega}_{\mathrm{right}}(\mathcal{W})\ge \tau_{\mathrm{hand}},\
\bar{\omega}(\mathcal{W})\ge \tau_{\mathrm{avg}},
\]
where $\tau_{\mathrm{hand}}$ and $\tau_{\mathrm{avg}}$ are fixed thresholds. In practice, we set $\tau_{\mathrm{hand}}=25$ and $\tau_{\mathrm{avg}}=30$.

\subsection{Dataset Metrics}
\label{sec:dataset_metrics}
In Table~\ref{tab:dataset_comparison_new}(\textbf{b}), we use several metrics to evaluate the quality of data interaction and action intensity. Specifically, we employ four key indicators.
\textbf{Contact Ratio} measures the fraction of frames where inter-hand contact occurs, representing the density of interaction.
\textbf{Contact Duration} denotes the average length of continuous contact spans, reflecting the stability of interactions.
\textbf{Contact Freq} counts the number of distinct contact events per minute, indicating the complexity and richness of the interaction. 
\textbf{Motion Intensity} quantifies the magnitude of fine-grained finger dynamics, as defined in Sec.~\ref{sec:Intensity Aware}.
\section{Additional Details on Motion Captioning}
\label{sec:rule-based feature extraction}

As Sec.~\ref{sec:3_auto_annotation} introduces, the objective of kinematic feature extraction is to transform raw hand motion sequences into structured, semantically meaningful representations that can be subsequently interpreted directly by an LLM. We compute six types of kinematic descriptors for bimanual hand motion. When the value of a kinematic descriptor exhibits changes over a period of time, we extract it as an ``event.'' These events collectively constitute the structured features of the two-hand motion.

\noindent \textbf{Kinematic Descriptors.} We define six types of kinematic descriptors to comprehensively characterize the motion: \textit{Finger Flexing}, \textit{Finger Spacing}, \textit{Finger-finger Distance}, \textit{Palm-palm Relation}, \textit{Finger-palm Distance}, and \textit{Wrist Trajectory}. These descriptors aim to comprehensively capture the motion by covering both local, single-hand features and global, bimanual dynamics.

\noindent \textit{Finger Flexing}. For a specific joint of a finger, let $\boldsymbol{p}\in \mathbb{R}^3$ denote the joint's coordinates, with $\boldsymbol{p}^{\mathrm{pre}}$ and $\boldsymbol{p}^{\mathrm{next}}$ representing the coordinates of the predecessor and successor joints, respectively. Define $\boldsymbol{v}^{\mathrm{pre}} = \boldsymbol{p}^{\mathrm{pre}} - \boldsymbol{p}$ and $\boldsymbol{v}^{\mathrm{next}} = \boldsymbol{p}^{\mathrm{next}} - \boldsymbol{p}$. We compute the ``signed angle'' $\theta$ of joint $\boldsymbol{p}$'s flexion using the dot and cross products of $\boldsymbol{v}^{\mathrm{pre}}$ and $\boldsymbol{v}^{\mathrm{next}}$.  If $\theta>0$, the joint exhibits normal flexion; if $\theta < 0$, the joint is in a hyperextended state.

\noindent \textit{Finger Spacing}. The finger spacing of two adjacent fingers is measured by the angle formed between their respective finger directions, computed from the MCP to PIP joints.

\noindent \textit{Finger-finger Distance}. The distance between any two fingertips represents \textit{Finger-finger Distance}. This includes both intra-hand distances between different fingertips of the same hand and inter-hand distances between left and right hand fingertips.

\noindent \textit{Palm-palm Relation}. For a single hand, we first randomly sample 100 points within the convex hull formed by the wrist joint and the five MCP joints, representing the palm's point cloud $\mathcal{G}^h = \{\boldsymbol{q}_i^h \in \mathbb{R}^3\}_{i=1}^{100}$ $(h=\mathrm{L}, \mathrm{R})$. From all point pairs between $\mathcal{G}^{\mathrm{L}}$ and $\mathcal{G}^{\mathrm{R}}$, we select the 30 closest pairs $\{(\boldsymbol{q}_i^{\mathrm{L}}, \boldsymbol{q}_i^{\mathrm{R}})\}_{i=1}^{30}$. Computing the differences of their world coordinates yields 30 vectors $\{\boldsymbol{v_i}\}_{i=1}^{30}$. We then average these vectors to obtain $\bar{\boldsymbol{v}} \in \mathbb{R}^3$, which serves as the kinematic descriptors for the left-to-right palm relation.

\noindent \textit{Finger-palm Distance}. For a fingertip joint $\boldsymbol{p}$ of one hand, we identify the 5 closest points $\{\boldsymbol{q}_i\}_{i=1}^5$ in the other hand's palm cloud $\mathcal{G}$. The average distance between $\boldsymbol{p}$ and these points serves as the this descriptor.

\noindent \textit{Wrist Trajectory}. We use the world coordinate of the wrist joint at the current timestep.

\noindent \textbf{Event Segmentation.} In general, if the value of a kinematic descriptor exhibits a detectable change over a time period, we extract it as an ``event.'' Additionally, if the feature's value remains essentially constant throughout the entire motion, we also consider this as a distinct type of event.

\begin{table}[h]
    \centering
    \caption{Quantitative state intervals for different hand motion features used in event segmentation.}
    \label{tab:feature_states_intervals}
    \resizebox{\columnwidth}{!}{%
    \begin{tabular}{|c|c|c|}
        \hline
        \textbf{Feature} & \textbf{Range} & \textbf{State} \\
        \hline
        \multirow{4}{*}{Finger Flexing ($^\circ$)} & $[-180, -20)$ & Hyper extend \\
        \cline{2-3}
        & $[-20, 30)$ & Fully extend \\
        \cline{2-3}
        & $[30, 60)$ & Partially bent \\
        \cline{2-3}
        & $[60, 180)$ & Fully bent \\
        \hline
        \multirow{2}{*}{Finger Spacing ($^\circ$)} & $[0, 20)$ & Closed \\
        \cline{2-3}
        & $[20, 180)$ & Open \\
        \hline
        \multirow{2}{*}{Finger-Finger Distance (m)} & $[0, 0.02)$ & Contact \\
        \cline{2-3}
        & $[0.02, +\infty)$ & No Contact \\
        \hline
        \multirow{3}{*}{Finger-Palm Distance (m)} & $[0, 0.025)$ & Contact \\
        \cline{2-3}
        & $[0.025, 0.035)$ & Near \\
        \cline{2-3}
        & $[0.035, +\infty)$ & Far \\
        \hline
    \end{tabular}}
\end{table}

As shown in Table \ref{tab:feature_states_intervals}, we have defined value intervals for certain descriptors, with each interval corresponding to a semantic label. For \textit{Finger Flexing}, \textit{Finger Spacing}, \textit{Finger-finger Distance}, and \textit{Finger-palm Distance}, events are characterized as transitions from one state to another or maintaining a constant state throughout the motion.

\begin{figure*}
    \centering
    \includegraphics[width=1\textwidth]{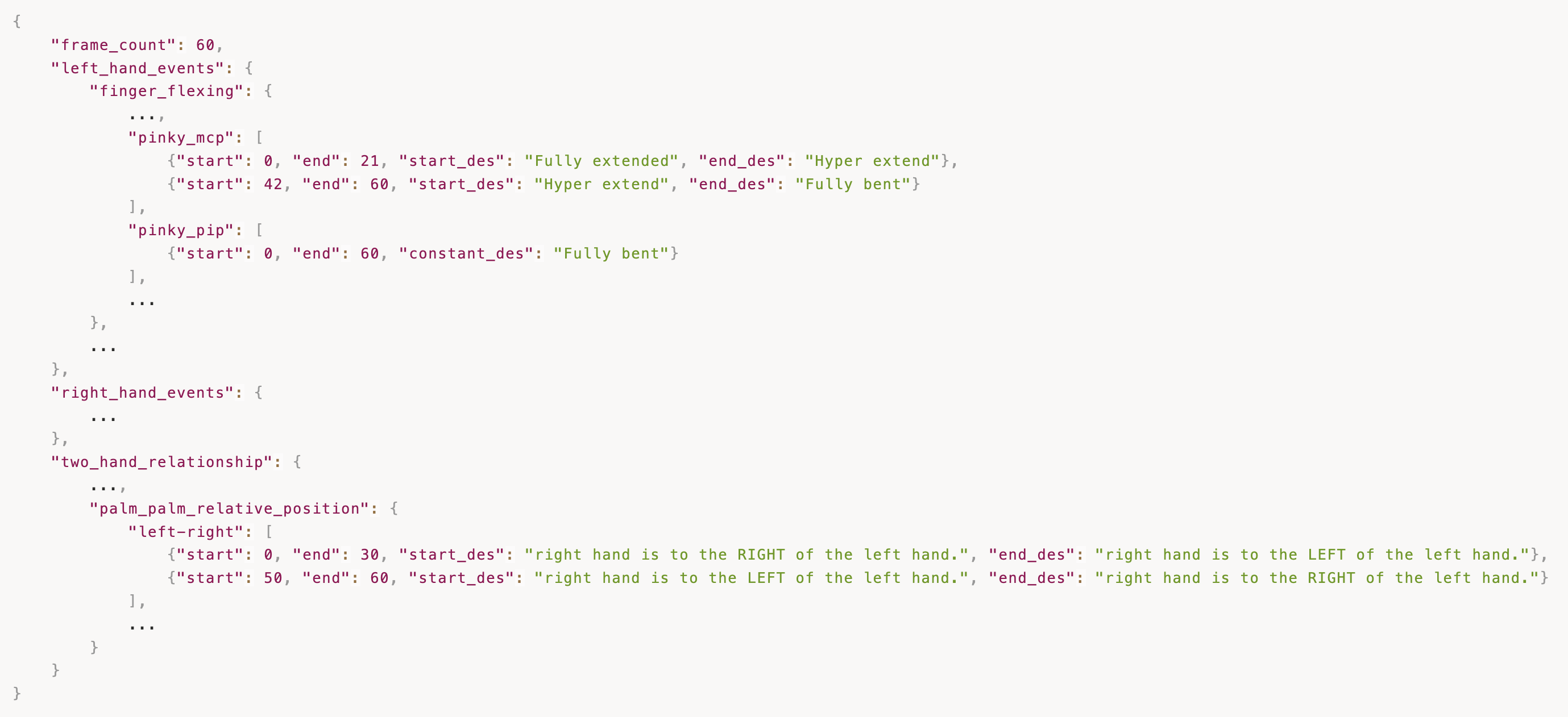}
    \caption{Structured representation of extracted motion features. The JSON format captures temporal segmentation with frame indices and semantic state descriptions for transition events and constant events.}
    \label{fig:feature_json_example}
\end{figure*}

For \textit{Wrist Trajectory} and \textit{Palm-palm Relation}, we decompose their vectors along the X, Y, and Z axes of the world coordinate system, where changes along these axes correspond to spatial movements in left-right, forward-backward, and up-down directions, respectively. Figure~\ref{fig:feature_json_example} shows an example of extracted feature.

\noindent\textbf{Prompt Design.}
\begin{figure*}[t]
  \centering
  \includegraphics[width=\textwidth]{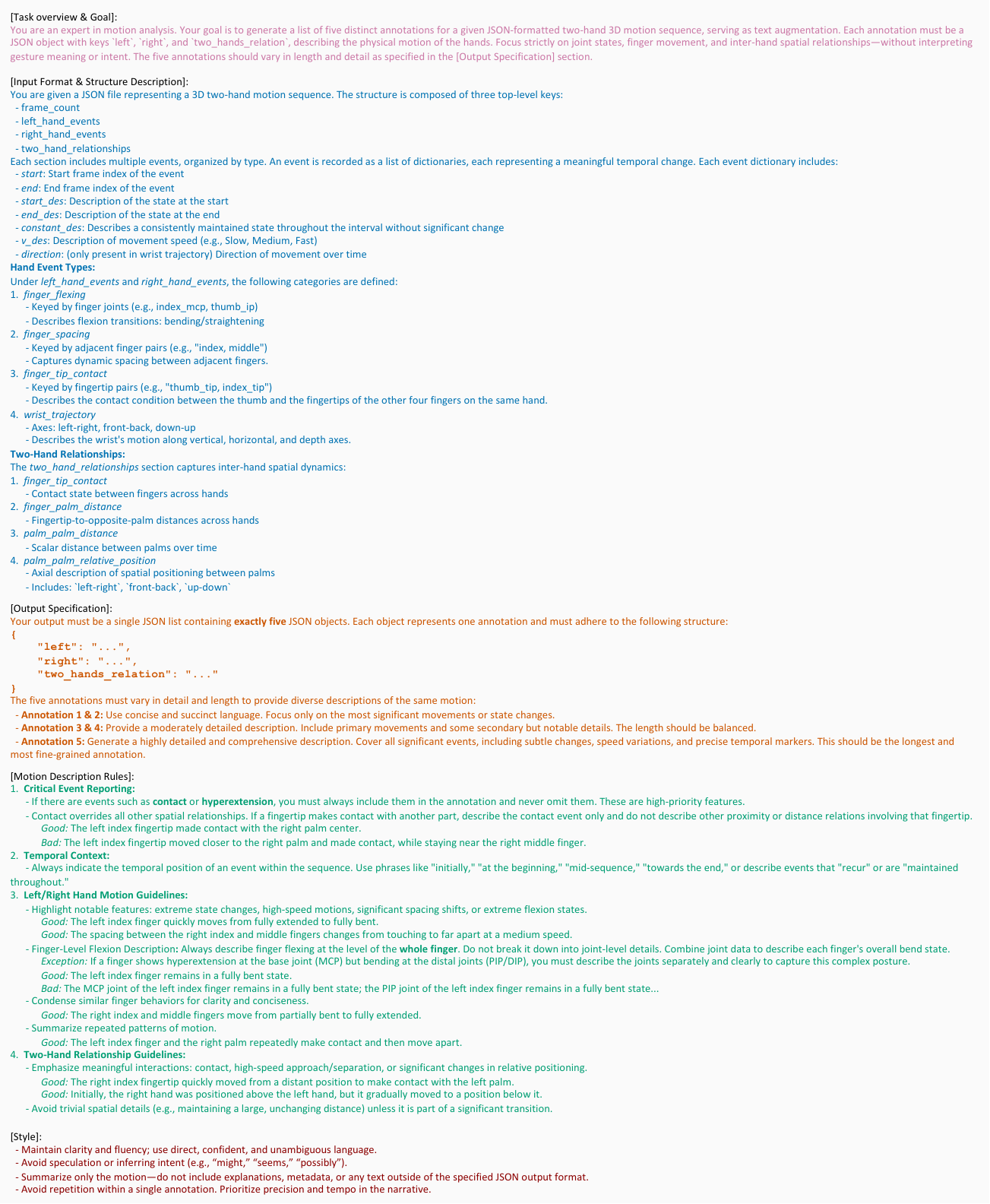} 
  \caption{Our prompt design to facilitate LLMs to summarize kinematic features. This is for bimanual motion captioning.}
  \label{fig:prompt}
\end{figure*}
Figure \ref{fig:prompt} illustrates our prompt design. By feeding the large language model JSON-formatted features, it can generate five distinct annotations at varying levels of granularity as requested.

\section{Hand Motion Representation}
\label{sec:diff_data}
We here provide the detailed definition of the representation component, rotational scalar. At frame $i$, let $\boldsymbol{v}^i \in \mathbb{R}^3$ denote the vector formed from the MCP joint of the little finger to the MCP joint of the index finger. For a given joint, let $\alpha$ be the angle formed by the joint itself, its predecessor joint, and its successor joint along the finger's kinematic chain. We project angle $\alpha$ onto the plane perpendicular to $\boldsymbol{v}^i$, and the resulting projected angle magnitude serves as the ``rotation scalar'' $s$ for this joint. We concatenate the 3D coordinates and rotation scalar for each joint at each frame

\begin{equation}
\boldsymbol{x}^i = [\boldsymbol{p}^i; \boldsymbol{s}^i] \in \mathbb{R}^{2J\times 4},
\end{equation}

\noindent where $[\cdot\,;\, \cdot]$ denotes the concatenation operation. Thus, $\boldsymbol{x} = \{\boldsymbol{x}^1, \boldsymbol{x}^2, \dots, \boldsymbol{x}^F\}\in\mathbb{R}^{F\times 2J\times 4}$ is our data representation.

\section{Versatile Bimanual Motion Generation}
\label{sec:vers}
\subsection{Masked Partial Denoising for Skeleton Control}
\label{sec:appendix_masked_pd}

Our versatile generation module is implemented using a masked
partial-denoising strategy inspired by~\cite{shafir2024human}.
The goal is to explicitly condition a subset, \eg, keyframes, wrist trajectories, or single-hand constraints,
while letting the remaining degrees of freedom be generated.

During inference, the diffusion model first denoises the current
noisy sample $\boldsymbol{x}_t$ to obtain a clean prediction $\boldsymbol{x}_0$, and then
re-applies the forward noising process to produce the next
timestep sample $\boldsymbol{x}_{t-1}$.
Let $\boldsymbol{x}_0^{\mathrm{pred}} = \mathcal{G}(\boldsymbol{x}_t, t, T)$ denote the clean motion
predicted by the model given timestep $t$ and text condition $T$,
and let $\boldsymbol{x}_0^{\mathrm{gt}}$ denote the target motion that provides
conditions (\eg, keyframe poses or reference trajectories).
We index frames by $i\in\{0,\dots,L{-}1\}$ and joints by
$j\in\{0,\dots,2J-1\}$, and write $\boldsymbol{x}_0(i,j)$ for the
state of joint $j$ at frame $i$, where $L$ denotes the generation length, and $J$ denotes the number of joints per hand.

\noindent\textbf{Soft Interpolation.}
To enable smooth temporal control, we apply constraints at the joint level and softly extend them from selected \textit{keyframes to nearby frames}. For each joint $j \in \{0,\dots,2J-1\}$, we define a set of center frames $\mathcal{C}_j \subseteq \{0,\dots,L-1\}$ where the target motion is most strongly enforced. This formulation applies to any control setting in which only a subset of frames is specified, \eg, \textit{motion in-betweening}, \textit{keyframe-based generation}, and \textit{long-horizon generation}, detailed in Sec.~\ref{sec:appendix_masked_pd_tasks}.

For each center frame $i \in \mathcal{C}_j$, we define a local temporal window
\begin{equation}
    \mathcal{T}(i) = \{\, t \in \{0,\dots,L-1\} \mid |t-i| \le k_{\mathrm{trans}} \,\},
\end{equation}
where $k_{\mathrm{trans}}$ controls the transition range. Within this window, the constraint weight decays linearly with temporal distance:
\begin{equation}
    \gamma_{t,j}^{(i)} =
    p_{\mathrm{hard}} -
    (p_{\mathrm{hard}} - p_{\mathrm{soft}})
    \frac{|t-i|}{k_{\mathrm{trans}}},
\end{equation}
where $p_{\mathrm{hard}}=0.85$, $p_{\mathrm{soft}}=0.10$, and $k_{\mathrm{trans}}=5$.

If multiple windows overlap, we use the strongest weight:
\begin{equation}
  \gamma_{t,j} =
  \begin{cases}
    \max\limits_{i \in \mathcal{C}_j,\; t \in \mathcal{T}(i)} \gamma_{t,j}^{(i)},
    & \text{if } \exists\, i \in \mathcal{C}_j \text{ s.t. } t \in \mathcal{T}(i),\\
    0, & \text{otherwise.}
  \end{cases}
\end{equation}
Thus, $\gamma_{t,j}=0$ means joint $j$ at frame $t$ is unconstrained.

We then update the clean signal by interpolating between the model prediction and the target:
\begin{equation}
\boldsymbol{x}_0(t,j)
=
(1-\gamma_{t,j})\,\boldsymbol{x}_0^{\mathrm{pred}}(t,j)
+
\gamma_{t,j}\,\boldsymbol{x}_0^{\mathrm{gt}}(t,j).
\label{eq:masked_clean}
\end{equation}
At the center frame $t=i \in \mathcal{C}_j$, the constraint is strongest, with $\gamma_{i,j}=p_{\mathrm{hard}}$.

Finally, we re-noise the modified $\boldsymbol{x}_0$ using the standard forward diffusion step:
\begin{equation}
\boldsymbol{x}_{t-1}
=
\sqrt{\bar{\alpha}_{t-1}}\,\boldsymbol{x}_0
+
\sqrt{1-\bar{\alpha}_{t-1}}\,\boldsymbol{\epsilon},
\label{eq:renoise}
\end{equation}
where $\boldsymbol{\epsilon} \sim \mathcal{N}(0,I)$ and $\bar{\alpha}_{t-1}$ follows the training noise schedule.

\subsection{Task-Specific Mask Construction}
\label{sec:appendix_masked_pd_tasks}
Based on the per-joint masking formulation in Sec.~\ref{sec:appendix_masked_pd}, we implement versatile generation tasks by configuring the center frame set $\mathcal{C}_j$ for different groups of joints. 
For clarity, we define $\mathcal{J}_{\mathrm{all}}$ as the set of all joints, and define specific subsets (\eg, $\mathcal{J}_\mathrm{wrist}$) for spatial control.

\noindent \textbf{Motion In-betweening.}
We constrain the start and end of the sequence for all joints. We set the target frame set $T_\mathrm{target} = \{0, \dots, K_\mathrm{inbet}-1\} \cup \{L-K_\mathrm{inbet}, \dots, L-1\}$, with $K_\mathrm{inbet}=5$. The joint masks are configured as:
\begin{equation}
    \mathcal{C}_j = T_\mathrm{target}, \quad \forall j \in \mathcal{J}_\mathrm{all}.
\end{equation}

\noindent \textbf{Keyframe-Based Generation.}
Given sparse keyframes at indices $T_\mathrm{key} = \{t_1, \dots, t_k\}$, we enforce these poses on the full skeleton:
\begin{equation}
    \mathcal{C}_j = T_\mathrm{key}, \quad \forall j \in \mathcal{J}_\mathrm{all}.
\end{equation}

\noindent \textbf{Wrist Trajectories Generation.}
We constrain only the wrist joints throughout the entire sequence, leaving fingers free. Let $T_\mathrm{seq} = \{0, \dots, L-1\}$. We set:
\begin{equation}
    \mathcal{C}_j = 
    \begin{cases} 
    T_\mathrm{seq}, & \text{if } j \in \mathcal{J}_\mathrm{wrist}, \\
    \varnothing, & \text{otherwise}.
    \end{cases}
\end{equation}
This explicitly controls the wrist positions while setting $\gamma_{t,j}=0$ for all other joints.

\noindent \textbf{Hand-Reaction Synthesis.}
One hand (\eg, left hand) is fully constrained as the condition. Let $\mathcal{J}_\mathrm{left}$ denote the joints of the left hand. We set:
\begin{equation}
    \mathcal{C}_j = 
    \begin{cases} 
    T_\mathrm{seq}, & \text{if } j \in \mathcal{J}_\mathrm{left}, \\
    \varnothing, & \text{if } j \in \mathcal{J}_\mathrm{right}.
    \end{cases}
\end{equation}

\noindent \textbf{Long Horizon Generation.}
We generate motion auto-regressively. We take the last $K_\mathrm{hor} + k_\mathrm{trans}$ frames of the preceding generated sequence as the ground truth to constrain the start of the current sequence. We configure the center frames as:
\begin{equation}
    \mathcal{C}_j = \{ 0, \dots, K_\mathrm{hor} - 1 \}, \quad \forall j \in \mathcal{J}_\mathrm{all}.
\end{equation}
where the temporal smoothing radius is set to $k_\mathrm{trans}$.

\begin{table}[!t]
    \centering
    \caption{\textbf{Ablation study} on diffusion model for analyzing scaling curves. R-Precision is evaluated with a batch size of 16.}
    \label{tab:denser_configuration}
    \resizebox{\columnwidth}{!}{%
    \begin{tabular}{@{}cccc c ccc c@{}}
        \toprule
        \multicolumn{4}{c}{\textbf{Configuration}} & 
        \multirow{2}{*}{\textbf{FLOPs (G)}} & 
        \multicolumn{3}{c}{\textbf{R-Precision}$^\uparrow$} & 
        \multirow{2}{*}{\textbf{FID}$^\downarrow$} \\
        \cmidrule(lr){1-4} \cmidrule(lr){6-8}
        L & d & H & FFN & & Top 1 & Top 2 & Top 3 & \\
        \midrule
        2 & 256 & 4 & 1,024 & 3.83 & $0.117^{\pm 0.072}$ & $0.210^{\pm 0.088}$ & $0.295^{\pm 0.091}$ & $5.342^{\pm 0.073}$ \\
        4 & 256 & 4 & 1,024 & 4.22 & $0.160^{\pm 0.085}$ & $0.264^{\pm 0.098}$ & $0.355^{\pm 0.107}$ & $2.705^{\pm 0.057}$ \\
        4 & 384 & 6 & 1,536 & 5.22 & $0.185^{\pm 0.087}$ & $0.298^{\pm 0.059}$ & $0.389^{\pm 0.092}$ & $2.053^{\pm 0.033}$ \\
        8 & 384 & 6 & 1,536 & 6.98 & $0.262^{\pm 0.101}$ & $0.407^{\pm 0.112}$ & $0.501^{\pm 0.115}$ & $1.959^{\pm 0.066}$ \\
        8 & 512 & 8 & 2,048 & 9.74 & $0.318^{\pm 0.117}$ & $0.463^{\pm 0.124}$ & $0.566^{\pm 0.124}$ & $1.830^{\pm 0.033}$ \\
        12 & 512 & 8 & 2,048 & 12.87 & $0.300^{\pm 0.109}$ & $0.437^{\pm 0.119}$ & $0.536^{\pm 0.116}$ & $2.492^{\pm 0.030}$ \\
        12 & 768 & 12 & 3,072 & 24.62 & ${\pmb{0.411}^{\pm \pmb{0.127}}}$ & ${\pmb{0.570}^{\pm \pmb{0.127}}}$ & ${\pmb{0.672}^{\pm \pmb{0.117}}}$ & ${\pmb{1.695}^{\pm \pmb{0.039}}}$ \\
        \bottomrule
    \end{tabular}}
    \vspace{-0.5em}
\end{table}

\section{Evaluation Details}
\label{sec:eval_details}
\subsection{Contact Metrics}

To comprehensively  evaluate the accuracy of two-hand interactions generation, we decompose the contact metrics into two categories: intra-hand and inter-hand contacts.
For intra-hand contacts, we assess whether the thumb fingertip makes contact with each of the other four fingertips within a single hand. A contact is registered when the distance between two fingertips falls below a predefined threshold at any frame in the sequence.

For inter-hand contacts, we measure the minimum distance between the closest point pair across the two hands. A contact is detected when this minimum distance is below the threshold.

We extract contact labels for both intra-hand and inter-hand categories from both ground truth and generated motion. We then compute true positives (TP), true negatives (TN), and false positives (FP), and report standard metrics. Intra-hand evaluation is presented in Table~\ref{tab:ablation_model_size}, the precision of inter-hand evaluation is shown in Table~\ref{tab:ablation_interhand}.
\begin{figure*}
    \centering
    \begin{subfigure}[t]{0.20\linewidth}
        \centering
        \includegraphics[width=\linewidth]{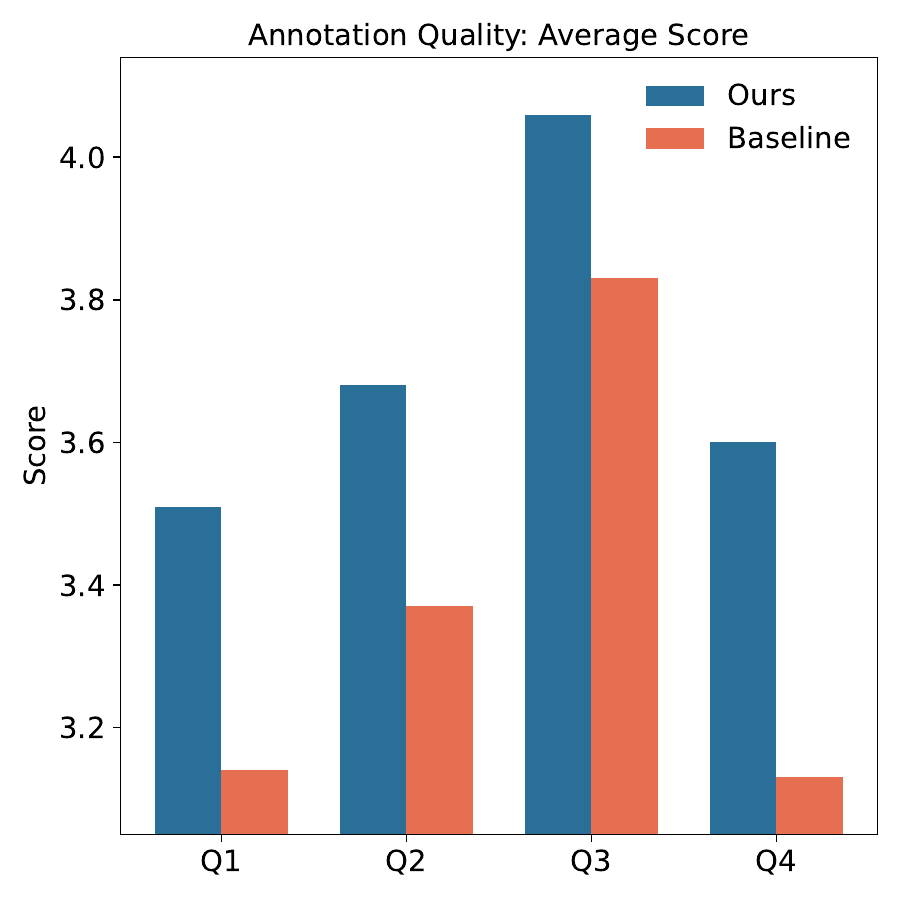}
        \caption{} 
        \label{fig:user study for annotation quality results}
    \end{subfigure}
    \hfill
    \begin{subfigure}[t]{0.20\linewidth}
        \centering
        \includegraphics[width=\linewidth]{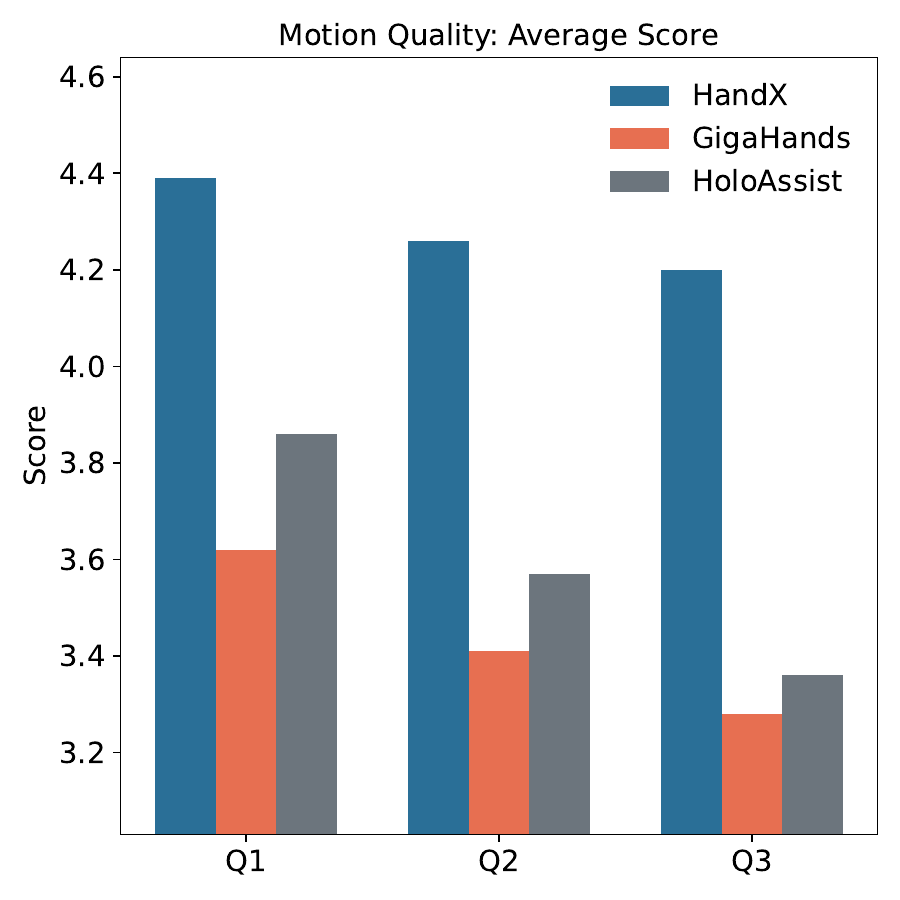}
        \caption{} 
        \label{fig:user study for motion quality results}
    \end{subfigure}
    \hfill
    \begin{subfigure}[t]{0.58\linewidth}
        \centering
        \includegraphics[width=\linewidth]{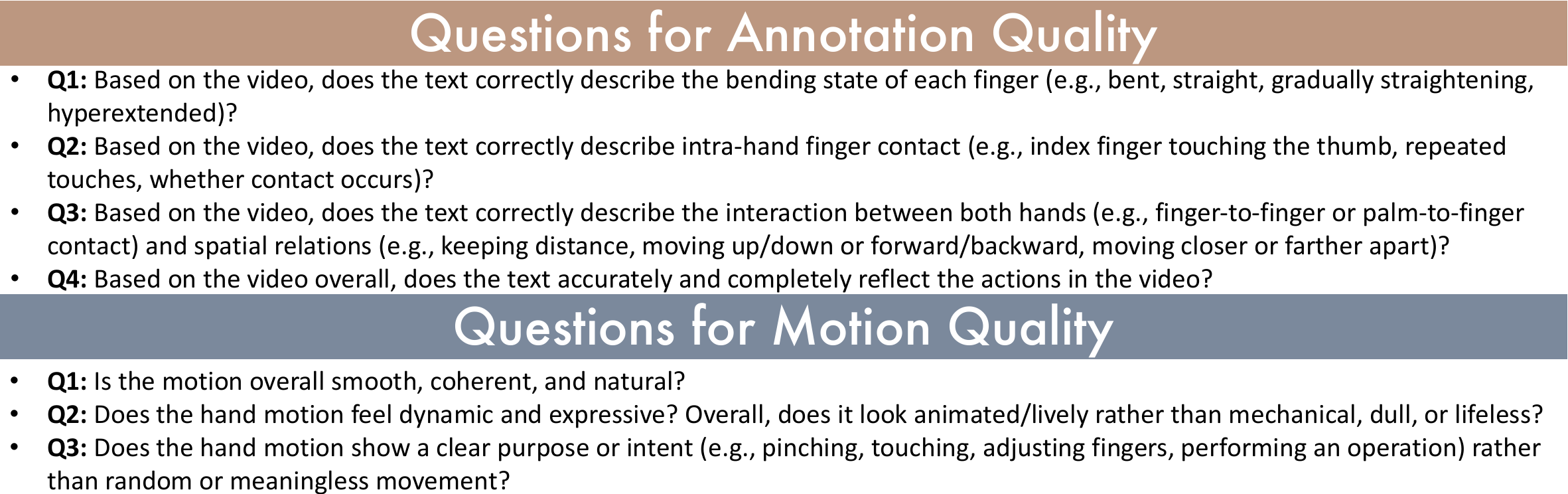}
        \caption{} 
        \label{fig:user study question design}
    \end{subfigure}

    \caption{\textbf{User study on data quality}. Our \oursx exhibits high-quality motion and annotations. For annotation quality evaluation, we compare our method with a baseline that utilizes Gemini 3 Pro~\cite{team2023gemini} to directly caption rendered motion videos.}
    \label{fig:user study for data quality}
\end{figure*}

\begin{table}
    \centering
        \caption{Detailed configurations of the diffusion and autoregressive models.
    Parameter counts exclude the frozen text encoder. Results from Figure~\ref{fig:fitting scaling equation} are extracted from this table.}
    \vspace{-0.5em}
    \resizebox{\columnwidth}{!}{%
    \begin{tabular}{ccccc}
        \hline
        Model & Layers & Latent dim & FF size & Trainable params \\
        \hline
        \multirow{4}{*}{Diffusion}
            & 4  & 256  & 1,024 & 4.63M  \\
            & 8  & 512  & 1,024 & 26.33M \\
            & 12 & 512  & 1,024 & 38.95M \\
            & 16 & 768 & 3,072 & 260.97M \\
        \midrule
        \multirow{3}{*}{Autoregressive}
            & 8  & 512  & 512  & 29.63M  \\
            & 12 & 768  & 768  & 92.27M \\
            & 16 & 1,024 & 1,024 & 215.31M \\
        \hline
    \end{tabular}
    }
    \label{tab:configuration_of_diffusion_models_AR}
\end{table}

\begin{table}[t]
    \centering
    \caption{\textbf{Ablation study on inter-hand interaction.} We report the inter-hand contact precision ($C_\mathrm{prec}$) on diffusion model across different dataset scales and model depths.}
    \label{tab:ablation_interhand}
    \resizebox{0.6\columnwidth}{!}{%
    \begin{tabular}{@{}ccc@{}}
        \toprule
        Dataset Ratio & Layers & Inter-hand $C_\mathrm{prec}$$^\uparrow$ \\
        \midrule
        0.05 & 4 & 0.7310 \\
        0.05 & 8 & 0.7381 \\
        0.05 & 12 & 0.7111 \\
        \midrule
        0.2 & 8 & 0.7066 \\
        0.2 & 12 & 0.7971 \\
        \midrule
        1.0 & 4 & 0.7551 \\
        1.0 & 8 & 0.7838 \\
        1.0 & 12 & \textbf{0.8593} \\
        \bottomrule
    \end{tabular}}
\end{table}

\subsection{Evaluator Details} \label{sec:evaluator_supp}
Our evaluator takes as input the global hand joint positions. Following~\cite{lu2023humantomato,petrovich2023tmr}, we jointly train the text encoder and the hand motion encoder. Unlike traditional approaches that rely on classification-based training~\cite{guo2022generating}, we adopt a sequence-level contrastive learning objective based on the InfoNCE loss~\cite{oord2018representation}. Given the text prompts $T = (T_L, T_R, T_I)$, we concatenate them with tokenizer-defined separator tokens to form the input sentence. The text encoder uses T5~\cite{raffel2020exploring} as its backbone.

\subsection{User Study on Data Quality}
To evaluate the quality of our data, we conduct a comprehensive user study focusing on two primary dimensions: \textbf{annotation quality} and \textbf{motion quality}. Regarding annotation quality, we compare our decoupled strategy against a baseline where Gemini 3 Pro~\cite{team2023gemini} is prompted to directly caption rendered videos. For motion quality, HandX is evaluated against GigaHands~\cite{fu2025gigahands} and HoloAssist~\cite{wang2023holoassist}. 

We recruit 20 participants for the study. Sequences are randomly permuted and distributed to participants. We ensure that each sequence receives ratings from at least three independent participants to maintain consistency. Following the question design shown in Figure~\ref{fig:user study question design}, participants rated the samples on a scale of 1 to 5, where higher scores indicate better quality. Figures~\ref{fig:user study for annotation quality results} and \ref{fig:user study for motion quality results} demonstrate that our approach significantly surpasses direct motion captioning in annotation quality and existing bimanual datasets in motion quality.

\subsection{User Study on Scaling Trend}
\label{sec:user_study}
To complement the quantitative scaling trends reported in the main paper, we conduct a perceptual user study to evaluate the visual quality and semantic consistency of the generated motion across different data scales.

\noindent\textbf{Experimental Setup.} We randomly sample 10 distinct text prompts from the test set to ensure an unbiased evaluation of the model's performance. For each prompt, we generate videos using the diffusion models trained on three different subsets of the training dataset: 5\%, 20\%, and 100\%, consistent with the quantitative ablation settings. All motions are rendered as 3D meshes. Crucially, to ensure clear visibility of the fine-grained finger interactions and spatial relations, we unify and optimize the camera viewing angles for each sequence.

\noindent\textbf{Procedure.} We recruit 10 participants to evaluate the generation quality; all of them are graduate or undergraduate students without previous knowledge on hand motion generation. For each of the 10 prompts, participants are presented with the text description and three generated videos displayed in a randomized order to prevent bias. Participants are allowed to replay the videos an unlimited number of times to inspect motion details carefully. We instruct participants to consider both motion naturalness and semantic alignment when making their choice. 

\noindent\textbf{Results.}
The user study shows a clear preference for the model trained on the full dataset. Specifically, the model trained on 100\% of the data receives 48\% of the votes, compared with 33\% for the model trained on 5\% of the data and 19\% for the model trained on 20\% of the data. This suggests that increasing the data scale leads to perceptually better motion quality and semantic alignment.

\section{License}
\label{sec:license}
Users must review and follow the original licenses for each sub-dataset utilized in \oursx. Please find the licenses of corresponding assets in the code directories, and below is a summary of the licenses for the assets we have used:

\begin{enumerate}
    \item {GigaHands}~\cite{fu2025gigahands} uses the Creative Commons Attribution-NonCommercial 4.0 International License.
    \item {HOT3D}~\cite{banerjee2025hot3d} uses the HOT3D Dataset License Agreement.
    \item {ARCTIC}~\cite{fan2023arctic} uses the Data \& Software Copyright License for non-commercial scientific research purposes.
    \item {H2O}~\cite{kwon2021h2o} uses custom Terms of Use restricted to academic and non-commercial purposes.
    \item 
    {HoloAssist}~\cite{wang2023holoassist} uses the CDLA v2 Permissive License.
\end{enumerate}
\section{Discussion}
\textbf{Limitations.}
Despite the comprehensive benchmark established in this work, several limitations remain. First, although \oursx significantly scales up bimanual motion data with fine-grained annotations, the dataset is still finite in both volume and diversity. Consequently, it cannot exhaustively cover the full spectrum of human dexterity or every possible interaction scenario found in the real world. Second, a portion of our training corpus is aggregated from existing public datasets. While we employ rigorous filtering and interpolation techniques to standardize such data, inherent quality issues in the raw sources, such as minor jitter or kinematic implausibility, cannot be completely eliminated.

\noindent
\textbf{Potential Negative Societal Impact.}
Our work focuses on generating realistic human hand motion, which has various positive applications. However, like other high-fidelity generative models, there is a risk of misuse in creating deepfakes or misleading content. The ability to synthesize dexterous hand motion from text could be used to fabricate videos of individuals performing actions they did not carry out. To mitigate this, we release our code and data under licenses that restrict usage to research and non-commercial purposes. Additionally, we address privacy concerns regarding the subjects involved in our data collection. We receive participants consent, and ensure the released dataset is strictly limited to skeletal motion representations, excluding any personally identifiable features to preserve anonymity.

%% file: main.bbl
\begin{thebibliography}{83}
\providecommand{\natexlab}[1]{#1}
\providecommand{\url}[1]{\texttt{#1}}
\expandafter\ifx\csname urlstyle\endcsname\relax
  \providecommand{\doi}[1]{doi: #1}\else
  \providecommand{\doi}{doi: \begingroup \urlstyle{rm}\Url}\fi

\bibitem[Abel et~al.(2024)Abel, Colotte, and Ouni]{abel2024towards}
Louis Abel, Vincent Colotte, and Slim Ouni.
\newblock Towards interpretable co-speech gestures synthesis using stargate.
\newblock In \emph{International Conference on Multimodal Interaction}, 2024.

\bibitem[Ahn et~al.(2018)Ahn, Ha, Choi, Yoo, and Oh]{ahn2018text2action}
Hyemin Ahn, Timothy Ha, Yunho Choi, Hwiyeon Yoo, and Songhwai Oh.
\newblock Text2action: Generative adversarial synthesis from language to action.
\newblock In \emph{ICRA}, 2018.

\bibitem[Ahuja and Morency(2019)]{ahuja2019language2pose}
Chaitanya Ahuja and Louis-Philippe Morency.
\newblock Language2pose: Natural language grounded pose forecasting.
\newblock In \emph{3DV}, 2019.

\bibitem[Baltatzis et~al.(2024)Baltatzis, Potamias, Ververas, Sun, Deng, and Zafeiriou]{baltatzis2024neural}
Vasileios Baltatzis, Rolandos~Alexandros Potamias, Evangelos Ververas, Guanxiong Sun, Jiankang Deng, and Stefanos Zafeiriou.
\newblock Neural sign actors: A diffusion model for 3d sign language production from text.
\newblock In \emph{CVPR}, 2024.

\bibitem[Banerjee et~al.(2025)Banerjee, Shkodrani, Moulon, Hampali, Han, Zhang, Zhang, Fountain, Miller, Basol, et~al.]{banerjee2025hot3d}
Prithviraj Banerjee, Sindi Shkodrani, Pierre Moulon, Shreyas Hampali, Shangchen Han, Fan Zhang, Linguang Zhang, Jade Fountain, Edward Miller, Selen Basol, et~al.
\newblock Hot3d: Hand and object tracking in 3d from egocentric multi-view videos.
\newblock In \emph{CVPR}, 2025.

\bibitem[Bensabath et~al.(2026)Bensabath, Petrovich, and Varol]{bensabath2025text}
L{\'e}ore Bensabath, Mathis Petrovich, and G{\"u}l Varol.
\newblock Text-driven 3d hand motion generation from sign language data.
\newblock \emph{CVPR}, 2026.

\bibitem[Bilge et~al.(2023)Bilge, Cinbis, and Ikizler-Cinbis]{9681230}
Y. Bilge, R. Cinbis, and N. Ikizler-Cinbis.
\newblock Towards zero-shot sign language recognition.
\newblock \emph{TPAMI}, 2023.

\bibitem[Bilge et~al.(2019)Bilge, Ikizler-Cinbis, and Cinbis]{bilge19zsslr}
Yunus~Can Bilge, Nazli Ikizler-Cinbis, and Ramazan~Gokberk Cinbis.
\newblock Zero-shot sign language recognition: Can textual data uncover sign languages?
\newblock In \emph{BMVC}, 2019.

\bibitem[Cha et~al.(2024)Cha, Kim, Yoon, and Baek]{cha2024text2hoi}
Junuk Cha, Jihyeon Kim, Jae~Shin Yoon, and Seungryul Baek.
\newblock Text2hoi: Text-guided 3d motion generation for hand-object interaction.
\newblock In \emph{CVPR}, 2024.

\bibitem[Chen et~al.(2024)Chen, Liu, Wang, Zeng, Li, and Chen]{chen2024diffsheg}
Junming Chen, Yunfei Liu, Jianan Wang, Ailing Zeng, Yu Li, and Qifeng Chen.
\newblock Diffsheg: A diffusion-based approach for real-time speech-driven holistic 3d expression and gesture generation.
\newblock In \emph{CVPR}, 2024.

\bibitem[Christen et~al.(2024)Christen, Hampali, Sener, Remelli, Hodan, Sauser, Ma, and Tekin]{christen2024diffh2o}
Sammy Christen, Shreyas Hampali, Fadime Sener, Edoardo Remelli, Tomas Hodan, Eric Sauser, Shugao Ma, and Bugra Tekin.
\newblock Diffh2o: Diffusion-based synthesis of hand-object interactions from textual descriptions.
\newblock In \emph{SIGGRAPH Asia}, 2024.

\bibitem[Delmas et~al.(2022)Delmas, Weinzaepfel, Lucas, Moreno-Noguer, and Rogez]{delmas2022posescript}
Ginger Delmas, Philippe Weinzaepfel, Thomas Lucas, Francesc Moreno-Noguer, and Gr{\'e}gory Rogez.
\newblock Posescript: 3d human poses from natural language.
\newblock In \emph{ECCV}, 2022.

\bibitem[Fan et~al.(2025)Fan, Lu, Dai, Yu, Xiao, Dou, Dong, Ma, and Wang]{fan2025go}
Ke Fan, Shunlin Lu, Minyue Dai, Runyi Yu, Lixing Xiao, Zhiyang Dou, Junting Dong, Lizhuang Ma, and Jingbo Wang.
\newblock Go to zero: Towards zero-shot motion generation with million-scale data.
\newblock In \emph{ICCV}, 2025.

\bibitem[Fan et~al.(2023)Fan, Taheri, Tzionas, Kocabas, Kaufmann, Black, and Hilliges]{fan2023arctic}
Zicong Fan, Omid Taheri, Dimitrios Tzionas, Muhammed Kocabas, Manuel Kaufmann, Michael~J Black, and Otmar Hilliges.
\newblock Arctic: A dataset for dexterous bimanual hand-object manipulation.
\newblock In \emph{CVPR}, 2023.

\bibitem[Fan et~al.(2024)Fan, Parelli, Kadoglou, Chen, Kocabas, Black, and Hilliges]{fan2024hold}
Zicong Fan, Maria Parelli, Maria~Eleni Kadoglou, Xu Chen, Muhammed Kocabas, Michael~J Black, and Otmar Hilliges.
\newblock Hold: Category-agnostic 3d reconstruction of interacting hands and objects from video.
\newblock In \emph{CVPR}, 2024.

\bibitem[Fang et~al.(2025)Fang, Chen, Wang, Zheng, Sui, and Tian]{fang2025signllm}
Sen Fang, Chen Chen, Lei Wang, Ce Zheng, Chunyu Sui, and Yapeng Tian.
\newblock Signllm: Sign language production large language models.
\newblock In \emph{ICCV}, 2025.

\bibitem[Ferstl et~al.(2019)Ferstl, Neff, and McDonnell]{ferstl2019multi}
Ylva Ferstl, Michael Neff, and Rachel McDonnell.
\newblock Multi-objective adversarial gesture generation.
\newblock In \emph{ACM SIGGRAPH Conference on Motion, Interaction and Games}, 2019.

\bibitem[Fu et~al.(2026)Fu, Wang, Qiao, Yang, Liu, and Zhao]{fu2026egograsp}
Hongming Fu, Wenjia Wang, Xiaozhen Qiao, Shuo Yang, Zheng Liu, and Bo Zhao.
\newblock Egograsp: World-space hand-object interaction estimation from egocentric videos.
\newblock \emph{arXiv preprint arXiv:2601.01050}, 2026.

\bibitem[Fu et~al.(2025)Fu, Zhang, Jiang, Fu, Funk, Ritchie, and Sridhar]{fu2025gigahands}
Rao Fu, Dingxi Zhang, Alex Jiang, Wanjia Fu, Austin Funk, Daniel Ritchie, and Srinath Sridhar.
\newblock Gigahands: A massive annotated dataset of bimanual hand activities.
\newblock In \emph{CVPR}, 2025.

\bibitem[Gjaci et~al.(2022)Gjaci, Recchiuto, and Sgorbissa]{gjaci2022towards}
Ariel Gjaci, Carmine~Tommaso Recchiuto, and Antonio Sgorbissa.
\newblock Towards culture-aware co-speech gestures for social robots.
\newblock \emph{International Journal of Social Robotics}, 14\penalty0 (6):\penalty0 1493--1506, 2022.

\bibitem[Guo et~al.(2022)Guo, Zou, Zuo, Wang, Ji, Li, and Cheng]{guo2022generating}
Chuan Guo, Shihao Zou, Xinxin Zuo, Sen Wang, Wei Ji, Xingyu Li, and Li Cheng.
\newblock Generating diverse and natural 3d human motions from text.
\newblock In \emph{CVPR}, 2022.

\bibitem[Ho et~al.(2020)Ho, Jain, and Abbeel]{ho2020denoising}
Jonathan Ho, Ajay Jain, and Pieter Abbeel.
\newblock Denoising diffusion probabilistic models.
\newblock \emph{NeurIPS}, 2020.

\bibitem[Hoque et~al.(2025)Hoque, Huang, Yoon, Sivapurapu, and Zhang]{hoque2025egodex}
Ryan Hoque, Peide Huang, David~J Yoon, Mouli Sivapurapu, and Jian Zhang.
\newblock Egodex: Learning dexterous manipulation from large-scale egocentric video.
\newblock \emph{arXiv preprint arXiv:2505.11709}, 2025.

\bibitem[Huang et~al.(2025)Huang, Chu, Tekin, Liang, Ma, Wang, Chen, Gleize, Xue, Lyu, et~al.]{huang2025hoigpt}
Mingzhen Huang, Fu-Jen Chu, Bugra Tekin, Kevin~J Liang, Haoyu Ma, Weiyao Wang, Xingyu Chen, Pierre Gleize, Hongfei Xue, Siwei Lyu, et~al.
\newblock Hoigpt: Learning long-sequence hand-object interaction with language models.
\newblock In \emph{CVPR}, 2025.

\bibitem[Jiang et~al.(2023)Jiang, Chen, Liu, Yu, Yu, and Chen]{jiang2023motiongpt}
Biao Jiang, Xin Chen, Wen Liu, Jingyi Yu, Gang Yu, and Tao Chen.
\newblock Motiongpt: Human motion as a foreign language.
\newblock \emph{NeurIPS}, 2023.

\bibitem[Kwon et~al.(2021)Kwon, Tekin, St{\"u}hmer, Bogo, and Pollefeys]{kwon2021h2o}
Taein Kwon, Bugra Tekin, Jan St{\"u}hmer, Federica Bogo, and Marc Pollefeys.
\newblock H2o: Two hands manipulating objects for first person interaction recognition.
\newblock In \emph{ICCV}, 2021.

\bibitem[Li et~al.(2025)Li, Christen, Wan, Cai, Liao, Sigal, and Ma]{li2025latenthoi}
Muchen Li, Sammy Christen, Chengde Wan, Yujun Cai, Renjie Liao, Leonid Sigal, and Shugao Ma.
\newblock Latenthoi: On the generalizable hand object motion generation with latent hand diffusion.
\newblock In \emph{CVPR}, 2025.

\bibitem[Lin et~al.(2023)Lin, Zeng, Lu, Cai, Zhang, Wang, and Zhang]{lin2023motion}
Jing Lin, Ailing Zeng, Shunlin Lu, Yuanhao Cai, Ruimao Zhang, Haoqian Wang, and Lei Zhang.
\newblock Motion-x: A large-scale 3d expressive whole-body human motion dataset.
\newblock \emph{NeurIPS}, 2023.

\bibitem[Lin(2025)]{lin2025handdiffuse}
Pei Lin.
\newblock Handdiffuse: generative controllers for two-hand interactions via diffusion models.
\newblock In \emph{AAAI}, 2025.

\bibitem[Liu et~al.(2024{\natexlab{a}})Liu, Zhu, Becherini, Peng, Su, Zhou, Zhe, Iwamoto, Zheng, and Black]{liu2024emage}
Haiyang Liu, Zihao Zhu, Giorgio Becherini, Yichen Peng, Mingyang Su, You Zhou, Xuefei Zhe, Naoya Iwamoto, Bo Zheng, and Michael~J Black.
\newblock Emage: Towards unified holistic co-speech gesture generation via expressive masked audio gesture modeling.
\newblock In \emph{CVPR}, 2024{\natexlab{a}}.

\bibitem[Liu et~al.(2025)Liu, Song, Huang, Liu, and Xu]{liu2025gesturelsm}
Pinxin Liu, Luchuan Song, Junhua Huang, Haiyang Liu, and Chenliang Xu.
\newblock Gesturelsm: Latent shortcut based co-speech gesture generation with spatial-temporal modeling.
\newblock In \emph{ICCV}, 2025.

\bibitem[Liu et~al.(2022{\natexlab{a}})Liu, Tripathi, Majumdar, and Wang]{liu2022joint}
Shaowei Liu, Subarna Tripathi, Somdeb Majumdar, and Xiaolong Wang.
\newblock Joint hand motion and interaction hotspots prediction from egocentric videos.
\newblock In \emph{CVPR}, 2022{\natexlab{a}}.

\bibitem[Liu et~al.(2022{\natexlab{b}})Liu, Liu, Jiang, Lyu, Wan, Shen, Liang, Fu, Wang, and Yi]{liu2022hoi4d}
Yunze Liu, Yun Liu, Che Jiang, Kangbo Lyu, Weikang Wan, Hao Shen, Boqiang Liang, Zhoujie Fu, He Wang, and Li Yi.
\newblock Hoi4d: A 4d egocentric dataset for category-level human-object interaction.
\newblock In \emph{CVPR}, 2022{\natexlab{b}}.

\bibitem[Liu et~al.(2024{\natexlab{b}})Liu, Yang, Si, Liu, Li, Zhang, Liu, and Yi]{liu2024taco}
Yun Liu, Haolin Yang, Xu Si, Ling Liu, Zipeng Li, Yuxiang Zhang, Yebin Liu, and Li Yi.
\newblock Taco: Benchmarking generalizable bimanual tool-action-object understanding.
\newblock In \emph{CVPR}, 2024{\natexlab{b}}.

\bibitem[Loper et~al.(2015)Loper, Mahmood, Romero, Pons-Moll, and Black]{SMPL:2015}
Matthew Loper, Naureen Mahmood, Javier Romero, Gerard Pons-Moll, and Michael~J Black.
\newblock {SMPL}: A skinned multi-person linear model.
\newblock \emph{ACM Transactions on Graphics}, 34\penalty0 (6), 2015.

\bibitem[Lu et~al.(2023)Lu, Chen, Zeng, Lin, Zhang, Zhang, and Shum]{lu2023humantomato}
Shunlin Lu, Ling-Hao Chen, Ailing Zeng, Jing Lin, Ruimao Zhang, Lei Zhang, and Heung-Yeung Shum.
\newblock {HumanTOMATO}: Text-aligned whole-body motion generation.
\newblock \emph{arXiv preprint arXiv:2310.12978}, 2023.

\bibitem[Lu et~al.(2025)Lu, Wang, Lu, Chen, Dai, Dong, Dou, Dai, and Zhang]{lu2025scamo}
Shunlin Lu, Jingbo Wang, Zeyu Lu, Ling-Hao Chen, Wenxun Dai, Junting Dong, Zhiyang Dou, Bo Dai, and Ruimao Zhang.
\newblock Scamo: Exploring the scaling law in autoregressive motion generation model.
\newblock In \emph{CVPR}, 2025.

\bibitem[Moon et~al.(2020)Moon, Yu, Wen, Shiratori, and Lee]{moon2020interhand2}
Gyeongsik Moon, Shoou-I Yu, He Wen, Takaaki Shiratori, and Kyoung~Mu Lee.
\newblock Interhand2.6m: A dataset and baseline for 3d interacting hand pose estimation from a single rgb image.
\newblock In \emph{ECCV}, 2020.

\bibitem[{NaturalPoint, Inc.}(2025)]{optitrack}
{NaturalPoint, Inc.}
\newblock Optitrack motion capture system.
\newblock \url{https://optitrack.com}, 2025.

\bibitem[Oord et~al.(2018)Oord, Li, and Vinyals]{oord2018representation}
Aaron van~den Oord, Yazhe Li, and Oriol Vinyals.
\newblock Representation learning with contrastive predictive coding.
\newblock \emph{arXiv preprint arXiv:1807.03748}, 2018.

\bibitem[Petrovich et~al.(2022)Petrovich, Black, and Varol]{petrovich22temos}
Mathis Petrovich, Michael~J. Black, and G{\"u}l Varol.
\newblock {TEMOS}: Generating diverse human motions from textual descriptions.
\newblock In \emph{ECCV}, 2022.

\bibitem[Petrovich et~al.(2023)Petrovich, Black, and Varol]{petrovich2023tmr}
Mathis Petrovich, Michael~J Black, and G{\"u}l Varol.
\newblock {TMR}: Text-to-motion retrieval using contrastive 3d human motion synthesis.
\newblock In \emph{ICCV}, 2023.

\bibitem[Prakash et~al.(2025)Prakash, Forsyth, and Gupta]{prakash2025bimanual}
Aditya Prakash, David Forsyth, and Saurabh Gupta.
\newblock Bimanual 3d hand motion and articulation forecasting in everyday images.
\newblock \emph{arXiv preprint arXiv:2510.06145}, 2025.

\bibitem[Raffel et~al.(2020)Raffel, Shazeer, Roberts, Lee, Narang, Matena, Zhou, Li, and Liu]{raffel2020exploring}
Colin Raffel, Noam Shazeer, Adam Roberts, Katherine Lee, Sharan Narang, Michael Matena, Yanqi Zhou, Wei Li, and Peter~J Liu.
\newblock Exploring the limits of transfer learning with a unified text-to-text transformer.
\newblock \emph{Journal of machine learning research}, 21\penalty0 (140):\penalty0 1--67, 2020.

\bibitem[Romero et~al.(2017)Romero, Tzionas, and Black]{MANO:SIGGRAPHASIA:2017}
Javier Romero, Dimitrios Tzionas, and Michael~J. Black.
\newblock Embodied hands: Modeling and capturing hands and bodies together.
\newblock \emph{ACM Transactions on Graphics}, 36\penalty0 (6), 2017.

\bibitem[Saleh(2022)]{saleh2022hybrid}
Khaled Saleh.
\newblock Hybrid seq2seq architecture for 3d co-speech gesture generation.
\newblock In \emph{International Conference on Multimodal Interaction}, 2022.

\bibitem[Shafir et~al.(2024)Shafir, Tevet, Kapon, and Bermano]{shafir2024human}
Yonatan Shafir, Guy Tevet, Roy Kapon, and Amit~H Bermano.
\newblock Human motion diffusion as a generative prior.
\newblock \emph{ICLR}, 2024.

\bibitem[Shao et~al.(2024)Shao, Pang, Zheng, Sun, and Liu]{shao2024human4dit}
Ruizhi Shao, Youxin Pang, Zerong Zheng, Jingxiang Sun, and Yebin Liu.
\newblock Human4dit: 360-degree human video generation with 4d diffusion transformer.
\newblock \emph{ACM Transactions on Graphics}, 43\penalty0 (6), 2024.

\bibitem[Taheri et~al.(2020)Taheri, Ghorbani, Black, and Tzionas]{taheri2020grab}
Omid Taheri, Nima Ghorbani, Michael~J Black, and Dimitrios Tzionas.
\newblock Grab: A dataset of whole-body human grasping of objects.
\newblock In \emph{ECCV}, 2020.

\bibitem[Taheri et~al.(2024)Taheri, Zhou, Tzionas, Zhou, Ceylan, Pirk, and Black]{taheri2024grip}
Omid Taheri, Yi Zhou, Dimitrios Tzionas, Yang Zhou, Duygu Ceylan, Soren Pirk, and Michael~J Black.
\newblock Grip: Generating interaction poses using spatial cues and latent consistency.
\newblock In \emph{3DV}, 2024.

\bibitem[Team et~al.(2023)Team, Anil, Borgeaud, Alayrac, Yu, Soricut, Schalkwyk, Dai, Hauth, Millican, et~al.]{team2023gemini}
Gemini Team, Rohan Anil, Sebastian Borgeaud, Jean-Baptiste Alayrac, Jiahui Yu, Radu Soricut, Johan Schalkwyk, Andrew~M Dai, Anja Hauth, Katie Millican, et~al.
\newblock Gemini: a family of highly capable multimodal models.
\newblock \emph{arXiv preprint arXiv:2312.11805}, 2023.

\bibitem[Tevet et~al.(2023)Tevet, Raab, Gordon, Shafir, Cohen-or, and Bermano]{tevet2022human}
Guy Tevet, Sigal Raab, Brian Gordon, Yoni Shafir, Daniel Cohen-or, and Amit~Haim Bermano.
\newblock Human motion diffusion model.
\newblock In \emph{ICLR}, 2023.

\bibitem[Thambiraja et~al.(2026)Thambiraja, Taheri, Danecek, Becherini, Pons-Moll, and Thies]{thambiraja2026clutch}
Balamurugan Thambiraja, Omid Taheri, Radek Danecek, Giorgio Becherini, Gerard Pons-Moll, and Justus Thies.
\newblock {CLUTCH}: Contextualized language model for unlocking text-conditioned hand motion modelling in the wild.
\newblock In \emph{ICLR}, 2026.

\bibitem[Wang et~al.(2024)Wang, Xu, Shi, Schumann, and Liu]{wang2024furelise}
Ruocheng Wang, Pei Xu, Haochen Shi, Elizabeth Schumann, and C~Karen Liu.
\newblock F{\"u}relise: Capturing and physically synthesizing hand motion of piano performance.
\newblock In \emph{SIGGRAPH Asia}, 2024.

\bibitem[Wang et~al.(2023)Wang, Kwon, Rad, Pan, Chakraborty, Andrist, Bohus, Feniello, Tekin, Frujeri, et~al.]{wang2023holoassist}
Xin Wang, Taein Kwon, Mahdi Rad, Bowen Pan, Ishani Chakraborty, Sean Andrist, Dan Bohus, Ashley Feniello, Bugra Tekin, Felipe~Vieira Frujeri, et~al.
\newblock Holoassist: an egocentric human interaction dataset for interactive ai assistants in the real world.
\newblock In \emph{ICCV}, 2023.

\bibitem[Wang et~al.(2026)Wang, Xu, Guo, Zhou, Gong, Wang, Wang, and Gui]{wang2026unleashing}
Ziyin Wang, Sirui Xu, Chuan Guo, Bing Zhou, Jiangshan Gong, Jian Wang, Yu-Xiong Wang, and Liangyan Gui.
\newblock Unleashing guidance without classifiers for human-object interaction animation.
\newblock In \emph{ICLR}, 2026.

\bibitem[Wen et~al.(2024)Wen, Pan, Ohkawa, Yang, Pan, Sato, Komura, and Wang]{wen2024generative}
Yilin Wen, Hao Pan, Takehiko Ohkawa, Lei Yang, Jia Pan, Yoichi Sato, Taku Komura, and Wenping Wang.
\newblock Generative hierarchical temporal transformer for hand pose and action modeling.
\newblock In \emph{ECCV}, 2024.

\bibitem[Xiao et~al.(2025)Xiao, Lu, Pi, Fan, Pan, Zhou, Feng, Zhou, Peng, and Wang]{xiao2025motionstreamer}
Lixing Xiao, Shunlin Lu, Huaijin Pi, Ke Fan, Liang Pan, Yueer Zhou, Ziyong Feng, Xiaowei Zhou, Sida Peng, and Jingbo Wang.
\newblock Motionstreamer: Streaming motion generation via diffusion-based autoregressive model in causal latent space.
\newblock \emph{arXiv preprint arXiv:2503.15451}, 2025.

\bibitem[Xu and Wang(2024)]{xu2024synchronize}
Pei Xu and Ruocheng Wang.
\newblock Synchronize dual hands for physics-based dexterous guitar playing.
\newblock In \emph{SIGGRAPH Asia}, 2024.

\bibitem[Xu et~al.(2023{\natexlab{a}})Xu, Li, Wang, and Gui]{xu2023interdiff}
Sirui Xu, Zhengyuan Li, Yu-Xiong Wang, and Liang-Yan Gui.
\newblock {InterDiff}: Generating 3d human-object interactions with physics-informed diffusion.
\newblock In \emph{ICCV}, 2023{\natexlab{a}}.

\bibitem[Xu et~al.(2023{\natexlab{b}})Xu, Wang, and Gui]{xu2023stochastic}
Sirui Xu, Yu-Xiong Wang, and Liangyan Gui.
\newblock Stochastic multi-person 3d motion forecasting.
\newblock In \emph{ICLR}, 2023{\natexlab{b}}.

\bibitem[Xu et~al.(2024)Xu, Wang, Wang, and Gui]{xu2024interdreamer}
Sirui Xu, Ziyin Wang, Yu-Xiong Wang, and Liang-Yan Gui.
\newblock {InterDreamer}: Zero-shot text to 3d dynamic human-object interaction.
\newblock In \emph{NeurIPS}, 2024.

\bibitem[Xu et~al.(2025{\natexlab{a}})Xu, Chao, Bian, Mousavian, Wang, Gui, and Yang]{xu2025dexplore}
Sirui Xu, Yu-Wei Chao, Liuyu Bian, Arsalan Mousavian, Yu-Xiong Wang, Liangyan Gui, and Wei Yang.
\newblock Dexplore: Scalable neural control for dexterous manipulation from reference scoped exploration.
\newblock In \emph{CoRL}, 2025{\natexlab{a}}.

\bibitem[Xu et~al.(2025{\natexlab{b}})Xu, Li, Zhang, Xu, Long, Wang, Lu, Dong, Jiang, Gupta, Wang, and Gui]{xu2025interact}
Sirui Xu, Dongting Li, Yucheng Zhang, Xiyan Xu, Qi Long, Ziyin Wang, Yunzhi Lu, Shuchang Dong, Hezi Jiang, Akshat Gupta, Yu-Xiong Wang, and Liang-Yan Gui.
\newblock Interact: Advancing large-scale versatile 3d human-object interaction generation.
\newblock In \emph{CVPR}, 2025{\natexlab{b}}.

\bibitem[Xu et~al.(2026)Xu, Schulter, Ziyadi, He, Fei, Wang, and Gui]{xu2026interprior}
Sirui Xu, Samuel Schulter, Morteza Ziyadi, Xialin He, Xiaohan Fei, Yu-Xiong Wang, and Liang-Yan Gui.
\newblock {InterPrior}: Scaling generative control for physics-based human-object interactions.
\newblock In \emph{CVPR}, 2026.

\bibitem[Xu et~al.(2025{\natexlab{c}})Xu, Xu, Wang, and Gui]{xu2025moreact}
Xiyan Xu, Sirui Xu, Yu-Xiong Wang, and Liang-Yan Gui.
\newblock {MoReact}: Generating reactive motion from textual descriptions.
\newblock \emph{arXiv preprint arXiv:2509.23911}, 2025{\natexlab{c}}.

\bibitem[Yang et~al.(2023)Yang, Wu, Li, Zhang, Hao, Bao, Cheng, and Xiao]{yang2023diffusestylegesture}
Sicheng Yang, Zhiyong Wu, Minglei Li, Zhensong Zhang, Lei Hao, Weihong Bao, Ming Cheng, and Long Xiao.
\newblock Diffusestylegesture: Stylized audio-driven co-speech gesture generation with diffusion models.
\newblock \emph{arXiv preprint arXiv:2305.04919}, 2023.

\bibitem[Yazdian et~al.(2023)Yazdian, Liu, Lagasse, Mohammadi, Cheng, and Lim]{yazdian2023motionscript}
Payam~Jome Yazdian, Eric Liu, Rachel Lagasse, Hamid Mohammadi, Li Cheng, and Angelica Lim.
\newblock Motionscript: Natural language descriptions for expressive 3d human motions.
\newblock \emph{arXiv preprint arXiv:2312.12634}, 2023.

\bibitem[Ye et~al.(2026)Ye, Li, Rong, and Liu]{ye2026whole}
Yufei Ye, Jiaman Li, Ryan Rong, and C~Karen Liu.
\newblock Whole: World-grounded hand-object lifted from egocentric videos.
\newblock \emph{arXiv preprint arXiv:2602.22209}, 2026.

\bibitem[Yu et~al.(2025)Yu, Zafeiriou, and Birdal]{yu2025dyn}
Zhengdi Yu, Stefanos Zafeiriou, and Tolga Birdal.
\newblock Dyn-hamr: Recovering 4d interacting hand motion from a dynamic camera.
\newblock In \emph{CVPR}, 2025.

\bibitem[Zhang et~al.(2023{\natexlab{a}})Zhang, Zhang, Cun, Zhang, Zhao, Lu, Shen, and Shan]{zhang2023generating}
Jianrong Zhang, Yangsong Zhang, Xiaodong Cun, Yong Zhang, Hongwei Zhao, Hongtao Lu, Xi Shen, and Ying Shan.
\newblock Generating human motion from textual descriptions with discrete representations.
\newblock In \emph{CVPR}, 2023{\natexlab{a}}.

\bibitem[Zhang et~al.(2025{\natexlab{a}})Zhang, Deng, Ma, and Potamias]{zhang2025hawor}
Jinglei Zhang, Jiankang Deng, Chao Ma, and Rolandos~Alexandros Potamias.
\newblock Hawor: World-space hand motion reconstruction from egocentric videos.
\newblock In \emph{CVPR}, 2025{\natexlab{a}}.

\bibitem[Zhang et~al.(2025{\natexlab{b}})Zhang, Zhang, An, Li, Zhang, Hu, and Liu]{zhang2025manidext}
Jiajun Zhang, Yuxiang Zhang, Liang An, Mengcheng Li, Hongwen Zhang, Zonghai Hu, and Yebin Liu.
\newblock Manidext: Hand-object manipulation synthesis via continuous correspondence embeddings and residual-guided diffusion.
\newblock \emph{TPAMI}, 2025{\natexlab{b}}.

\bibitem[Zhang et~al.(2022)Zhang, Cai, Pan, Hong, Guo, Yang, and Liu]{zhang2022motiondiffuse}
Mingyuan Zhang, Zhongang Cai, Liang Pan, Fangzhou Hong, Xinying Guo, Lei Yang, and Ziwei Liu.
\newblock Motiondiffuse: Text-driven human motion generation with diffusion model.
\newblock \emph{arXiv preprint arXiv:2208.15001}, 2022.

\bibitem[Zhang et~al.(2023{\natexlab{b}})Zhang, Guo, Pan, Cai, Hong, Li, Yang, and Liu]{zhang2023remodiffuse}
Mingyuan Zhang, Xinying Guo, Liang Pan, Zhongang Cai, Fangzhou Hong, Huirong Li, Lei Yang, and Ziwei Liu.
\newblock Remodiffuse: Retrieval-augmented motion diffusion model.
\newblock In \emph{CVPR}, 2023{\natexlab{b}}.

\bibitem[Zhang et~al.(2024)Zhang, Huang, Zhou, Zhang, Yu, Wang, and Xu]{zhang2024both2hands}
Wenqian Zhang, Molin Huang, Yuxuan Zhou, Juze Zhang, Jingyi Yu, Jingya Wang, and Lan Xu.
\newblock Both2hands: Inferring 3d hands from both text prompts and body dynamics.
\newblock In \emph{CVPR}, 2024.

\bibitem[Zhang et~al.(2025{\natexlab{c}})Zhang, Dabral, Golyanik, Choutas, Alvarado, Beeler, Habermann, and Theobalt]{zhang2025bimart}
Wanyue Zhang, Rishabh Dabral, Vladislav Golyanik, Vasileios Choutas, Eduardo Alvarado, Thabo Beeler, Marc Habermann, and Christian Theobalt.
\newblock Bimart: A unified approach for the synthesis of 3d bimanual interaction with articulated objects.
\newblock In \emph{CVPR}, 2025{\natexlab{c}}.

\bibitem[Zhang et~al.(2025{\natexlab{d}})Zhang, Shi, Yang, Ni, Ye, and Wang]{zhang2025openhoi}
Zhenhao Zhang, Ye Shi, Lingxiao Yang, Suting Ni, Qi Ye, and Jingya Wang.
\newblock Openhoi: Open-world hand-object interaction synthesis with multimodal large language model.
\newblock In \emph{NeurIPS}, 2025{\natexlab{d}}.

\bibitem[Zhou et~al.(2022)Zhou, Bhatnagar, Lenssen, and Pons-Moll]{zhou2022toch}
Keyang Zhou, Bharat~Lal Bhatnagar, Jan~Eric Lenssen, and Gerard Pons-Moll.
\newblock Toch: Spatio-temporal object-to-hand correspondence for motion refinement.
\newblock In \emph{ECCV}, 2022.

\bibitem[Zhou et~al.(2024)Zhou, Bhatnagar, Lenssen, and Pons-Moll]{zhou2024gears}
Keyang Zhou, Bharat~Lal Bhatnagar, Jan~Eric Lenssen, and Gerard Pons-Moll.
\newblock Gears: Local geometry-aware hand-object interaction synthesis.
\newblock In \emph{CVPR}, 2024.

\bibitem[Zhu et~al.(2023)Zhu, Liu, Liu, Qian, Liu, and Yu]{zhu2023taming}
Lingting Zhu, Xian Liu, Xuanyu Liu, Rui Qian, Ziwei Liu, and Lequan Yu.
\newblock Taming diffusion models for audio-driven co-speech gesture generation.
\newblock In \emph{CVPR}, 2023.

\bibitem[Zou et~al.(2024)Zou, Yuan, Du, Wang, Liu, Xu, Chen, and Ji]{zou2024parco}
Qiran Zou, Shangyuan Yuan, Shian Du, Yu Wang, Chang Liu, Yi Xu, Jie Chen, and Xiangyang Ji.
\newblock Parco: Part-coordinating text-to-motion synthesis.
\newblock In \emph{ECCV}, 2024.

\bibitem[Zuo et~al.(2025)Zuo, Potamias, Ververas, Deng, and Zafeiriou]{zuo2025signs}
Ronglai Zuo, Rolandos~Alexandros Potamias, Evangelos Ververas, Jiankang Deng, and Stefanos Zafeiriou.
\newblock Signs as tokens: A retrieval-enhanced multilingual sign language generator.
\newblock In \emph{ICCV}, 2025.

\end{thebibliography}
